\documentclass[11pt]{article}

\usepackage[margin=1in]{geometry}
\usepackage{amsmath,amssymb,amsthm}
\usepackage{booktabs}
\usepackage{graphicx}
\usepackage{tikz}
\usetikzlibrary{positioning,arrows.meta,shapes.geometric}
\usepackage[hidelinks]{hyperref}
\usepackage{caption}
\usepackage{float}

\newtheorem{definition}{Definition}

\title{Auditability Is Not One Property: Rule Overlap, Behavioural Agreement, and Composition in Reinforcement Learning}
\author{Liu Hung Ming\thanks{PARRAWA AI, \texttt{cyril.liu@gmail.com} }}

\begin{document}
\maketitle

% ----------------------------------------------------------------------
\begin{abstract}
Reinforcement learning (RL) policies are typically delivered as opaque neural checkpoints, and the
logs of a training run verify that training happened rather than explaining what the policy learned.
We study whether a population of independently trained policy networks can be described and composed
at the level of discrete behavioral rules, and under which conditions that composition is
trustworthy. Our first contribution is a decomposition rather than a metric: we treat
\emph{auditability} as six separately testable predicates---trace integrity, lossless coding, rule
coverage, behavioural agreement, composition quality, and value-model reliability---and show that
passing one does not license another. This paper measures each predicate on its own terms, states
every prediction before seeing any result, and reports the protocol's failure-driven design history
as part of the record.

We introduce a protocol in which (i)~agents share one frozen, calibration-derived symbolizer;
(ii)~a passive observer converts each agent's logged transitions into a bank of confidence-scored
condition--action rules; (iii)~an append-only hash-bound ledger and exact replay verifier certify,
relative to the recorded artifact and environment version, that the recorded trace is the trace
executed; and (iv)~rules from multiple seeds are fused offline by a confidence-ranked arbitration
policy with an explicit fallback for blind spots.

Three findings constrain what such a description layer can claim. First, symbolic overlap does not
transfer to behavioral agreement: across seeds the rule sets overlap substantially, yet on fresh
states the independently trained policies agree at a rate indistinguishable from chance. A high
score on the coverage predicate is therefore not evidence for the behavioural-agreement predicate,
which is the paper's central negative result. Second, arbitration is not generation: the fused policy
selects among rules that already exist, so its correctness is bounded by how the bank was built
rather than by the knowledge in it. Third, we show that the apparent failure of this protocol on a
conflict-dominated task is an induction/deployment protocol artifact rather than a task limitation:
the bank was induced from sampled training actions but evaluated under argmax, and under that
mismatch a confidence-ranked arbiter loses to a random arbiter; rebuilding the bank under a protocol
consistent with deployment reverses the ordering. We also contribute negative evidence that a
fitted-Q generalized-policy-improvement baseline collapses in both environments, which places strict
limits on claiming that rule-level fusion is better than value-based composition.

The claims are bounded deliberately. The one composition comparison that favours rule-level fusion
rests on a comparator selected after training by a criterion we show to be weak, on a task whose
ceiling is reached by half the population, and on a fusion policy that remains far below the best
held-out actor; we report it as exploratory rather than as a declared win. Cross-hardware replay is
untested, the mechanism behind the conflict-regime reversal remains undecomposed, and the
communication result is bounded to a designed task.
\end{abstract}

% ----------------------------------------------------------------------
\section{Introduction}
\label{sec:intro}

Reinforcement learning systems are increasingly deployed where their decisions matter, and a natural
question accompanies every deployment: \emph{what, exactly, has the trained agent learned to do?}
For a policy network, the honest answer usually requires replaying checkpoints and reading log
statistics. Both are useful, and both stop short of explanation: a checkpoint is an opaque function
from observations to actions, and a training log is evidence that a procedure ran, not a description
of the behavior it produced.

\subsection{Auditability is not a single property}
\label{sec:intro:props}

The word \emph{auditability} is convenient and dangerous: it suggests a single grade that a system
either earns or does not. Our study instead treats it as six disjoint predicates, each with its own
estimand and its own negative result. Collapsing them into one number would hide the fact that the
same system can pass some and fail others.
This decomposition follows a broader auditability discipline: questions about whether coordination
logic can be reconstructed from discrete-symbol logs without hidden channels can likewise be split
into separately testable predicates whose verdicts may disagree \cite{liu2025aim}. We use the same
discipline here for single-policy description: each of our six predicates has its own estimand, and
passing one does not license another.

\begin{table}[t]
\centering\small
\caption{Auditability disaggregated into six separately measured predicates. The right column is a
boundary each predicate enforces; none of them licenses the others.}
\label{tab:props}
\begin{tabular}{@{}p{0.22\linewidth}p{0.34\linewidth}p{0.34\linewidth}@{}}
\toprule
Predicate & Estimand & What it does \emph{not} establish \\
\midrule
Trace integrity & hash chain + exact replay & causal explanation \\
Lossless coding & exact reconstruction of the trace & policy explanation; storage win \\
Rule coverage & held-out coverage of induced rules & correctness of the policy \\
Behavioural agreement & fresh-state action agreement & a shared policy core \\
Composition quality & return + decision mix & optimal arbitration \\
Value-model reliability & held-out value diagnostics & valid GPI composition \\
\bottomrule
\end{tabular}
\end{table}

We distinguish six \emph{audit predicates} from three \emph{verification mechanisms}. The predicates
are trace integrity, lossless coding, rule coverage, behavioural agreement, composition quality, and
value-model reliability. Hash-chain integrity, environment replay, and provenance resolution are
verification mechanisms: they supply evidence for selected predicates, but they are not themselves
predicates and are not interchangeable with one another. Keeping the two categories apart is what
prevents a passing mechanism check from being reported as a passing predicate.

This decomposition is the spine of the paper. Section~\ref{sec:prelim} states the predicates formally;
Section~\ref{sec:evolution} shows which design intuitions failed each one; and
Section~\ref{sec:experiments} reports a pass/fail verdict per predicate. The central negative result
is that \emph{rule overlap} (a coverage predicate) carries no information about \emph{behavioural
agreement} (a different predicate), so a high score on one must never be read as a high score on the
other.

This gap matters most when several trained agents are available and we want to combine them. If a
population of independently initialized policies each mastered the same task in a different way, can
we compare their behaviors, identify what they agree on, and assemble a single policy that inherits
the agreement? Three obstacles appear immediately. First, two policies with identical returns may
behave differently on the same states, so return statistics cannot stand in for behavioral
descriptions. Second, if we discretize states in a way that is private to each agent, the resulting
descriptions are not comparable across agents. Third, using induced rules to \emph{constrain} a
policy during training is known---and, as we show, empirically---to distort the very behavior we
hope to describe: an overly restrictive action mask trades exploration for compliance and can return
a suboptimal policy \cite{stappert2026integrating}, masking alters the policy gradient rather than
merely restricting the action set \cite{huang2022masking}, and injecting external knowledge into the
exploration process changes what the learner is able to discover \cite{garcia2015comprehensive}.

Our objective in this paper is therefore not another method for squeezing out higher returns. It is
to turn trained policies into auditable, composable artifacts, and to measure precisely where that
transformation succeeds and where it breaks. We pursue a rule-level description: a policy is
reported as a bank of condition--action rules over a \emph{shared} discretized state space, together
with an execution log whose fidelity is operationally verified and that ties every recorded decision
to the exact environment transition
that produced it.

The path from that intuition to a working protocol is a sequence of failures, and we treat those
failures as evidence rather than obstacles. Early attempts at grammar-based compression of behavior
turned out to be lossy; using induced rules to constrain training improved some seeds and damaged
others; a high-confidence rule admission threshold produced policies with 100\% blind spots;
compression of the audit log increased its size; and a contaminated ledger schema had to be repaired
by redefining the hashable core of a training record. Each failure changed the protocol, and each
change is documented with its cause.

The main protocol that survived this process has four components. A frozen shared state symbolizer
maps every agent's observations into a common discrete vocabulary. A passive rule observer converts
logged transitions into confidence-scored rules. A replay verifier and a hash-bound ledger certify
that the logs are faithful. And an offline fusion policy arbitrates the rule banks of multiple seeds
under an explicit blind-spot fallback. We evaluate the protocol on two control benchmarks with eight
seeds each.

Our contributions are bounded and deliberately conservative. (1) Under the shared symbolizer,
rule-level fusion of eight independently trained actors exceeded the selected single-actor comparator
on the matched CartPole-v1 evaluation set by $+42.31$ return points (95\% bootstrap interval
$[23.98,61.93]$) while remaining auditable at the artifact level (replay and provenance checks pass
under the evaluated setup); the comparator was selected post hoc by training return and the task
exhibits ceiling effects, so we report this comparison as exploratory rather than as a
pre-declared win. We make the sensitivity of that headline explicit: on the same matched episodes four
of the eight individual actors reach the ceiling (500.0) and fusion is $294.04$ points below the best
held-out actor, so the result is a statement about the selection rule rather than about fusion being a
strong single policy. (2) On Acrobot-v1 the same mechanism appears to fail in an instructive way: conflicts dominate and
confidence-ranked arbitration underperforms random arbitration under the original sampled-bank
induction. We show this is an induction/deployment protocol artifact, not a task limitation: re-inducing
the bank from argmax-consistent (greedy) labels recovers fusion to $-141.77$ and reverses the ordering,
while the previously cited 22.5\% Monte-Carlo agreement probe is a 2-actor continuation that does not
transfer to the 8-actor fusion (A2).
(3) We contribute negative evidence that constrains the whole approach: rule-set overlap does not
imply behavioral agreement (51.4\% agreement on fresh states despite a pairwise Jaccard overlap of
0.56), and a fitted-Q GPI baseline collapses in both environments, which places strict limits on
claiming that rule fusion is ``better than value-based composition.'' (4) We report an
occupancy-independent opportunity baseline that bounds what description-layer composition can buy.
Coverage as usually reported is measured on the steps a policy itself visits and therefore rises when
a policy is driven into the region its own bank dominates; over the eight \emph{frozen} banks the
observed condition vocabulary is $193$ of the $256$ grid cells on CartPole-v1 ($75.4\%$) and $608$ of
$4096$ on Acrobot-v1 ($14.8\%$), of which a single bank already accepts $59.7\%$ and $34.8\%$
respectively. The region where a description-layer arbiter has two or more admissible candidates---the
intersection of all eight banks---is $38$ conditions on CartPole-v1 and $2$ on Acrobot-v1, and on
Acrobot-v1 none of those two receives the same action from all eight. Merging banks therefore adds
descriptive reach slowly, and symbol-level agreement between banks ($0.999$ mean on CartPole-v1) is
shown to be a quantizer artifact with no discriminative power rather than evidence of a shared policy
core.

Both tasks are evaluated with the \emph{same} provenance-preserving rule-fusion framework
(Section~\ref{sec:method:fusion}); the divergences reported below are task-specific---action
spaces, actor populations, rule-bank construction protocols, and arbitration diagnostics---not
differences of fusion method.

The paper proceeds from the simple question to the full protocol. Section~\ref{sec:prelim} fixes
notation and states the audit problem. Section~\ref{sec:method} describes the final protocol.
Section~\ref{sec:evolution} reconstructs the failure-driven design history that produced it.
Section~\ref{sec:experiments} reports the experiments, including all negative results.
Sections~\ref{sec:related}--\ref{sec:conclusion} position the work, discuss it, and conclude.

% ----------------------------------------------------------------------
\section{Preliminaries and Problem Statement}
\label{sec:prelim}

\subsection{Markov decision process and agents}
\label{sec:prelim:mdp}

We consider a finite-horizon Markov decision process (MDP) $(\mathcal{S},\mathcal{A},P,r,\gamma)$
with state space $\mathcal{S}\subseteq\mathbb{R}^{d}$, discrete action space
$\mathcal{A}=\{1,\ldots,K\}$, transition kernel $P$, bounded reward $r$, and discount
$\gamma\in(0,1]$. A policy $\pi\colon \mathcal{S}\to\Delta(\mathcal{A})$ maps states to action
distributions. The discounted return of a trajectory is
\begin{equation}
G_t \;=\; \sum_{\tau=t}^{T} \gamma^{\tau-t}\, r(s_\tau,a_\tau),
\label{eq:return}
\end{equation}
and the value of a policy is its expected return from the initial state distribution.

We train agents with REINFORCE \cite{williams1992simple}, the policy-gradient update
\begin{equation}
\nabla J(\theta) \;=\; \mathbb{E}_{\tau\sim\pi_\theta}\Big[\big(G_t - b(s_t)\big)\,\nabla_\theta \log \pi_\theta(a_t\mid s_t)\Big],
\label{eq:reinforce}
\end{equation}
where $\theta$ parameterizes the policy network and $b(s_t)$ is a learned state-value baseline.
Each agent is a two-hidden-layer network with 64 units per layer; eight independent seeds are used throughout.

\subsection{The audit problem}
\label{sec:prelim:audit}

Given a trained deterministic policy $\pi$ (the argmax of a network), an auditor can observe its
behavior by executing it in the environment and recording transitions
$(s,a,r,s',\mathrm{term},\mathrm{trunc})$. We call the resulting artifact a \emph{trace}. Three
questions define our problem:

\begin{enumerate}
  \item \emph{Fidelity.} Can the auditor certify, without trusting the training harness, that the
        trace is exactly what was executed, and that a checkpoint reproduces the logged decisions?
  \item \emph{Description.} Can the trace be compressed into a finite, inspectable set of rules such
        that the rules are comparable across agents that were trained independently?
  \item \emph{Composition.} Can rule banks from several agents be fused into one policy, and under
        which conditions does the fused policy inherit the strengths---rather than the
        disagreements---of its sources?
\end{enumerate}

\paragraph{Scope distinction.}
The three questions above are frequently conflated with stronger claims they do not support.
Re-executing recorded transitions verifies the record. It does not recover policy weights over time,
optimizer state, gradients, or the random-number state that produced them, and it cannot establish
that the recorder was honest. We therefore keep four evidentiary layers strictly separate:
\emph{trace fidelity} (the record is what ran), \emph{policy description} (the rule bank is a finite
summary), \emph{behavioural agreement} (two policies act alike on fresh states), \emph{provenance}
(which rule and source produced a decision), and \emph{causal explanation} (why a decision was made).
A result at one layer is never cited as evidence at another.

\subsection{Scope and non-claims}
\label{sec:prelim:nonclaims}

To pre-empt over-reading, we state up front what this study does \emph{not} claim. We do not claim
general or complete auditability of arbitrary RL training; that symbolic overlap entails behavioural
consensus; that induced rules are human-readable explanations of a neural policy; a storage-compression
advantage over generic compression; optimal or mechanism-explained arbitration; validated value-model
composition; or a natural-language semantics for the discrete communication symbols. Every positive
claim below is bounded to two benchmark tasks, eight seeds per task, one shared quantizer per task, and
one software/hardware configuration.

Throughout, we distinguish three evidentiary layers: \emph{identification-bearing} measurements
(behavioral agreement on fresh states), \emph{diagnostic} measurements (rule statistics, fitted-Q
diagnostics), and \emph{operational-health} checks (hash-chain validity, replay equality). A perfect
score on an operational-health check certifies the artifact; it does not certify the policy.

% ----------------------------------------------------------------------
\section{Related Work}
\label{sec:related}

We organize the positioning around three axes, following the most rigorous framing in the literature:
\emph{(A)~assumption}---what each related line assumes about the policy or the environment;
\emph{(B)~demonstrated achievement}---what it actually shows; \emph{(C)~conditions of failure}---where
its guarantee breaks; and \emph{(D)~relationship to the present work}. This structure is adopted
deliberately so that each related method is judged by the same four questions we ask of our own
protocol.

\emph{Policy description and interpretability of RL.} Post-hoc interpretability of policies
typically produces saliency or attribution maps rather than inspectable behavioral rules, and such
maps do not certify behavior on unseen states. Rule extraction for RL has a long line of work using
decision trees and program synthesis. A closer line of work extracts structured policies from neural
controllers: VIPER uses imitation-style policy extraction guided by a neural policy and its value
estimates, with formal verification results on several control tasks \cite{bastani2018viper}, and
other work directly optimizes interpretable decision-tree policies or controls the
performance--complexity trade-off during policy learning \cite{roth2019conservative}. Our setting
differs in objective: we do not claim that the rule bank is a behaviorally equivalent replacement for
the neural policy; instead we evaluate trace anchoring, cross-seed comparability, provenance, and the
failure of rule overlap to predict behavior. The present study also differs in that rules are induced
from a \emph{shared, frozen symbolizer} so that rule banks from different seeds are directly
comparable, and in that every rule is anchored to a replayed, hash-bound trace.

\emph{Value-based composition.} Policy improvement theorems \cite{sutton2018reinforcement} show how
a value function can certify improvement over a base policy, and generalized policy improvement
(GPI) combines several value functions; formal GPI composes policies through approximate action-value
functions and provides a performance bound that depends on value-estimation error
\cite{barreto2017successor}. Our rule-level arbiter is \emph{not} an instance of theorem-backed GPI: it
ranks candidate actions by empirical rule confidence and support rather than by estimates of
source-policy action values $Q^{\pi_i}$, so no GPI-style policy-improvement guarantee applies to rule
fusion. We therefore use ``GPI baseline'' only for the separate fitted-Q comparator (\ref{eq:gpi}),
and we make no GPI-style policy-improvement claim for rule fusion. That comparator fails precisely
because the fitted critics are nearly constant, a failure mode documented for FQE under off-policy
coverage limits \cite{le2019batch,voloshin2021empirical} and consistent with the broader offline-RL
problem of distribution shift and extrapolation error: a critic trained from a fixed replay
distribution can assign unreliable values to state--action pairs induced by the composed policy
\cite{fujimoto2019offpolicy,kumar2020cql}. We read that failure as a validity diagnostic for this
implementation, not as evidence against value-based composition in general. We do not position rule
fusion as a replacement for value-based composition; the evidence supports only the bounded claim
that, under a shared symbolizer, rule-level arbitration can exceed the selected single-actor
comparator in a low-conflict regime.

\emph{Discrete symbolic communication and auditability.} Emergent communication can
appear under cooperative objectives \cite{foerster2016,havrylov2017}, but measurement is fraught:
Lowe et al.\ \cite{lowe2019} catalog pitfalls in which apparent communication is an artifact of the
objective or the analysis. The ``AI Mother Tongue'' line shows VQ-VAE-derived discrete symbols can
serve as an endogenous channel \cite{liu2025aim}, and its audit agenda asks whether coordination logic
can be reconstructed from discrete-symbol logs without hidden channels. The present work shares the
audit agenda but does not rely on emergent communication: our symbols are a fixed calibration-derived
abstraction, not a learned codebook, so the two lines are complementary rather than competing. We do
not claim any result about learned symbol systems. Our causal-symbol test
(Section~\ref{sec:experiments:causal}) is a positive demonstration that the causal question \emph{can}
be answered by intervention in a designed task, complementing Lowe et al.'s catalog of ways correlation
misleads.

\emph{Auditability of training records.} Ledgering and replay verification of training runs are
reproducibility practices. Our contribution on this axis is limited but explicit: a hash-bound
ledger over canonical payloads plus exact environment replay, with the negative findings that
human-readable serialization inflates size and that grammar-style codecs can fail to compress
traces. These are practical reproducibility findings, not new cryptography.

% ----------------------------------------------------------------------

\section{Method}
\label{sec:method}

\begin{figure}[t]
\centering
\begin{tikzpicture}[
  node distance=0.45cm and 0.9cm,
  box/.style={draw, rounded corners=2pt, align=center, minimum height=0.66cm, font=\small, fill=white, text width=2.35cm},
  arrow/.style={-{Stealth[length=2.2mm]}, thick},
  dim/.style={fill=black!6}
]
\node[box] (env) {Environment\\ (fixed reset seeds)};
\node[box, right=of env] (actor) {Policy network\\ $\pi_i$, seed $i$};
\node[box, dim, below=of actor] (sym) {Shared symbolizer $\varphi$\\ (frozen, calibrated)};
\node[box, below=of sym] (obs) {Rule observer\\ (support, confidence, mean reward)};
\node[box, below=of obs] (bank) {Rule bank $\mathcal{R}_i$};
\node[box, below=of bank] (fusion) {Offline fusion $F$\\ (confidence-ranked,\\ blind-spot fallback)};
\node[box, below=of fusion] (policy) {Fused policy $F(s)$};
\node[box, dim, right=of bank] (ledger) {Trace ledger\\ (hash-bound)};
\node[box, dim, below=of ledger] (replay) {Replay verifier};

\draw[arrow] (env) -- (actor);
\draw[arrow] (actor) -- (sym);
\draw[arrow] (sym) -- (obs);
\draw[arrow] (obs) -- (bank);
\draw[arrow] (bank) -- (fusion);
\draw[arrow] (fusion) -- (policy);
\draw[arrow] (actor.south) |- (ledger.west);
\draw[arrow] (ledger) -- (replay);
\draw[arrow,dashed] (replay.east) to[out=0,in=0,looseness=1.35] node[below,pos=0.5,font=\footnotesize]{certifies} (env.east);
\end{tikzpicture}
\caption{The auditable rule-induction protocol. A frozen shared symbolizer makes rule banks
comparable across seeds; a hash-bound ledger and replay verifier certify fidelity; offline fusion
composes rule banks into a single auditable policy. Dashed edge: the replay verifier re-executes
the recorded actions in the environment under the recorded reset seeds.}
\label{fig:pipeline}
\end{figure}
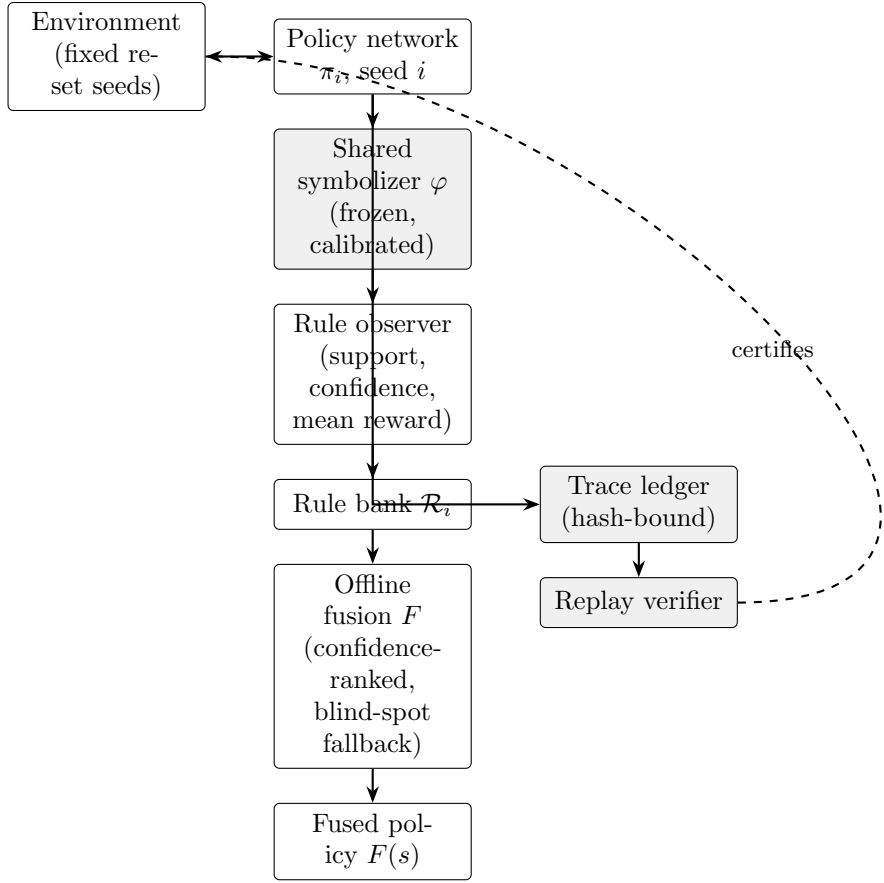

\subsection{Shared state symbolizer}
\label{sec:method:symbolizer}

All agents operate on the same continuous state space, but independently trained networks may rely
on different parts of it. Discrete symbolic abstractions can provide compact, behaviorally relevant
codes, and learned symbols can also serve as endogenous communication channels
\cite{vandenoord2017neural,liu2025aim}. Our audit setting deliberately does not rely on a learned
codebook: to make descriptions comparable, we fix \emph{one} symbolizer before any agent is trained
and freeze it. This makes determinism, replayability, and hashability explicit by construction;
a learned codebook could in principle provide the same properties, but only if its assignments and
parameters were separately preserved and bound to the audit record. For each state dimension
$j\in\{1,\ldots,d\}$, pooled empirical
quantiles are computed from separate random-policy roll-ins (24 episodes per seed), yielding three
cut points per dimension and four bins. The symbolizer is
\begin{equation}
\varphi(s) \;=\; \big(\,\mathrm{bin}_1(s_1),\ldots,\mathrm{bin}_d(s_d)\big)\;\in\;\{0,1,2,3\}^{d},
\label{eq:symbolizer}
\end{equation}
so the shared vocabulary has $4^{d}$ symbols ($256$ symbols for CartPole-v1, $4096$ for
Acrobot-v1). The quantiles are computed once and never updated, so every seed and every arm of the
experiment sees the same abstraction. This design choice is coupled to everything downstream; we
discuss its alternatives in Section~\ref{sec:discussion}.

The calibration roll-ins are generated by a uniformly random policy, not by any trained actor, and
their reset seeds are disjoint from both the training reset-seed range ($10^{4}\cdot\text{seed}+i$)
and the evaluation reset-seed range ($2\times10^{7}+i$); the calibration offset is $10^{5}$, so no
calibration episode is ever a training or evaluation episode, although the same eight seed
\emph{identities} are reused for the roll-in streams. Quantile cut points are computed separately per
environment, before any rule bank is constructed, and are never recomputed from held-out states. We
treat the four-bin resolution as a design choice rather than an optimized representation: no bin-count
search was performed, and the symbolizer is therefore a plausible confounder for every downstream
result, as Section~\ref{sec:discussion:limitations} states.

\begin{definition}[State symbolization]
\label{def:symbolizer}
A \emph{state symbolizer} $\varphi$ maps a real state to a tuple of bin indices
\begin{equation}
\varphi(s) \;=\; \big(\,\mathrm{bucketize}_{d}\!\big(s^{(d)};\,\theta_{d}\big)\,\big)_{d=1}^{D},
\label{eq:symbolize:def}
\end{equation}
where each $\theta_{d}$ holds the interior empirical-quantile edges of dimension $d$ over $B=4$
bins. The edges are fitted once from pooled random-policy calibration rollouts and then \emph{frozen},
so that all training seeds and arms within a task share the same symbol alphabet.
\end{definition}

\noindent Three consequences follow and are load-bearing. \emph{(i)} The symbolizer is built from pooled
calibration data; no seed fits its own discretizer. \emph{(ii)} Shared labels are \emph{not} shared
semantics: when each agent fits its own state discretizer, identical bin labels need not denote
identical raw-state regions---the largest observed edge difference under per-seed fitting was
$0.157$ observation units. \emph{(iii)} Symbol overlap is therefore \emph{not} behavioural equivalence.
The fix that makes cross-seed comparison possible is the single frozen symbolizer above.

\subsection{Rule bank and admission}
\label{sec:method:rules}

A \emph{rule} is a pair $(\mathbf{c},a)$ where $\mathbf{c}=\varphi(s)$ is a condition (a discrete
state symbol) and $a\in\mathcal{A}$ is an action. From the logged trace of agent $i$, the observer
computes, for every condition--action pair,
\begin{equation}
n_{\mathbf{c},a} \;=\; \#\{(s,a)\in \mathcal{T}_i : \varphi(s)=\mathbf{c}\},
\qquad
f_{\mathbf{c},a} \;=\; \frac{n_{\mathbf{c},a}}{\sum_{a'} n_{\mathbf{c},a'}},
\qquad
\mu_{\mathbf{c},a} \;=\; \frac{1}{n_{\mathbf{c},a}}\sum_{r\in \mathcal{T}_i(\mathbf{c},a)} r,
\label{eq:rulestats}
\end{equation}
where $\mathcal{T}_i$ is the transition set of agent $i$ and $r$ is the immediate reward. The
condition--action pair is admitted into the rule bank $\mathcal{R}_i$ if and only if
\begin{equation}
\mathrm{Adm}(\mathbf{c},a)\;\equiv\;\big(n_{\mathbf{c},a}\ge \tau_{s}\big)\ \wedge\ \big(f_{\mathbf{c},a}\ge \tau_{c}\big),
\qquad \tau_{s}=8,\ \tau_{c}=0.70 .
\label{eq:admission}
\end{equation}

\begin{definition}[Admitted state--action rule]
\label{def:rule}
Let $N(c,a)$ be the count of observed transitions whose symbol is $c=\varphi(s)$ and action is $a$,
and $N(c)=\sum_{a}N(c,a)$. The induced action for condition $c$ is
$a^{\star}(c)=\arg\max_{a}N(c,a)$ (ties to the smallest index), with support
$\kappa_{s}(c)=N(c,a^{\star})$ and confidence
\begin{equation}
\kappa(c) \;=\; \frac{N(c,a^{\star}(c))}{N(c)},
\qquad
\text{rule } c\!\rightarrow\!a^{\star}(c) \text{ is admitted iff }
\kappa_{s}(c)\ge \kappa_{s}^{\min}\ \wedge\ \kappa(c)\ge \tau .
\label{eq:rule:def}
\end{equation}
The admitted rules form a rule bank $\mathcal{R}$. Each rule stores the multiset of source step
identifiers that produced it, and the digest of that multiset.
\end{definition}

\noindent Equation~\eqref{eq:rule:def} uses a single normalization for confidence; support and confidence
are reported separately from coverage, so that a precise-but-rare rule is never described as broadly
applicable. The thresholds are not arbitrary. A high-confidence variant ($\tau_c=0.90$) of the
\emph{online} rule-constraint pilot admitted so few rules that fusion deferred to the fallback on
100\% of held-out decisions, and strict admission produced a 100\% blind-spot collapse
(Section~\ref{sec:evolution:threshold}); at $\tau_c=0.70$ with $\tau_s=8$ and an explicit blind-spot
fallback, coverage failure becomes observable rather than silent. High confidence is therefore neither
safety nor generalizability: rule admission is a \emph{passive} observation step and must never re-enter
the training loop.

We call $f_{\mathbf{c},a}$ the \emph{confidence} of the rule. The admission predicate is a
deterministic reduction of the trace: it compresses logged statistics without sampling, so its
output is exactly reproducible from the same trace.

\subsection{Audit ledger and replay verification}
\label{sec:method:ledger}

Fidelity is certified by two independent mechanisms.

\emph{Environment replay.} For every recorded transition, the verifier re-executes the environment
under the recorded reset seed with the recorded action and compares the resulting state, reward,
termination, and truncation flags against the recorded values:
\begin{equation}
\mathrm{Replay}\big((s,a,r,s',\mathrm{term},\mathrm{trunc});\,k\big)\ \equiv\ \big[\mathrm{Env}(s,a;k) = (s',r,\mathrm{term},\mathrm{trunc})\big],
\label{eq:replay}
\end{equation}
where $\mathrm{Env}(\cdot;k)$ is the deterministic environment transition under reset seed $k$ and
equality is exact float32 equality. Replay is a transport-level check: it certifies that the logged
trace is the executed trace, and it says nothing about whether the policy is good.

\emph{Hash-bound ledger.} Every training event (environment transition and optimizer update) is
serialized into a canonical payload and appended to a hash chain
\begin{equation}
h_0 = H(\mathrm{init}), \qquad h_{t} = H\big(h_{t-1} \,\Vert\, \mathrm{payload}_t\big),
\label{eq:hashchain}
\end{equation}
with SHA-256. The chain is recomputed from the raw payloads and compared against the recorded chain;
a mismatch at any position invalidates the entire run. An update ledger records, for each optimizer
step, the checkpoint hash before and after the step, so the chain certifies the sequence of weight
states as well as the sequence of transitions.

\paragraph{Three distinct evidence types.}
The audit ledger supplies three separable proofs that must not be merged into one ``audit score''.
\emph{(i)~Hash-chain integrity} certifies the record was not edited after the fact. \emph{(ii)~Environment
replay} certifies that each recorded transition can be regenerated by the same environment under the
same reset seed (a transport-level check). \emph{(iii)~Provenance resolution} certifies that a given
decision can be parsed back to the rule, source bank, arbiter outcome, and fallback that produced it.
The provenance predicate is
\begin{equation}
V_{\mathrm{prov}}(t) \;=\; \big[\,\mathrm{ref}_t \in \mathcal{R}_{i(t)} \;\wedge\;
\mathrm{act}(\mathrm{ref}_t)=a_t \;\wedge\; \mathrm{stat}(\mathrm{ref}_t)=\mathrm{stat}_t\,\big],
\label{eq:vprov}
\end{equation}
which an independent verifier evaluates for every decision in the fused trace.

\paragraph{The sense in which we use ``certified''.}
Because these mechanisms are easy to over-read, we fix their meaning here. We use ``certified'' only in
the operational sense just defined: a certification claim in this paper is relative to the recorded
artifact, the environment version, the reset seeds, and the verifier implementation. It is not a
certificate of the original training process, of the training harness configuration, of optimizer or
random-number state, or of the neural policy's internal mechanism. Hash-chain integrity and exact
replay establish that the stored record was not altered after the fact and that it can be regenerated
under the recorded environment and reset seeds; they do not establish that the record was produced by
the training procedure it claims, nor that the induced rules are a faithful semantic explanation of
the network. Where the paper says a layer is ``verified'', that is the claim being made and no more.

\paragraph{The hashable core.}
A subtle failure mode is a hash that covers non-core fields. In an early schema a sidecar counter and a
wrapper label entered the update hash even though they did not affect optimisation, so two runs with
identical optimizer behaviour hashed differently. The fix is a \emph{normalized core-update hash} that
excludes metadata (the previous hash, the record hash, and the grammar-generation annotation) and
covers only optimisation-relevant tensors---step index, parameter-gradient hash, and pre/post
checkpoint hashes. Equivalent runs then hash identically, and optimizer-equivalent arms (baseline vs.
audit-only) are demonstrably byte-identical.

\begin{definition}[Replay-and-provenance audit]
\label{def:audit}
A fused trace $\tau=(\rho_1,\ldots,\rho_T)$ \emph{passes} the replay-and-provenance audit,
written $\mathrm{PA}(\tau)=\mathrm{true}$, if and only if all three conditions hold:
\emph{(i)~environment replay passes} for every step (the replayed transition equals the logged
transition under the recorded reset seed, Eq.~\ref{eq:replay}); \emph{(ii)~provenance resolves}
for every decision (the cited rule reference belongs to the declared source bank, its action equals
the emitted action, and its support and confidence meet the admission thresholds, Eq.~\ref{eq:vprov});
and \emph{(iii)~the hash chain recomputes} exactly (Eq.~\ref{eq:hashchain}). The predicate is defined
only over the audited traces under the specified experiment and artifact scope---the eight-seed,
two-environment, shared-symbolizer study reported here---and asserts nothing about fused traces
outside that scope, nor that the training-time logger was honest.
\end{definition}

\subsection{Lossless trajectory codec}
\label{sec:method:codec}

Beyond the rule bank, every trace is also compressed by a \emph{lossless} codec so that a replayed
record can be checked against its compressed form. The codec is a Re-Pair-style pair-substitution
grammar with a bounded rule count ($\le 96$ rules), and decoding is asserted to be exact (the decoded
payload hashes to the input hash). Compression is measured as a two-part code length: the grammar
description plus the encoded sequence, using Elias-gamma counts for terminals and a fixed-width symbol
reference for non-terminals; the provenance bytes are reported separately. Writing the two-part
description length as
\begin{equation}
L_2(\tau) \;=\; L(\mathcal{G}_\tau) \;+\; L(\tau \mid \mathcal{G}_\tau),
\label{eq:twopart_code}
\end{equation}
where $\mathcal{G}_\tau$ is the induced grammar and $L(\tau\mid\mathcal{G}_\tau)$ is the length of the
trace encoded under it, we stress that $\mathcal{G}_\tau$ is a \emph{lossless trace grammar}: it
reconstructs the observed sequence exactly, but it is neither the state--action rule bank nor an
explanation of the neural policy. We call $L_2$ a \emph{two-part description length}, not a minimum
description length: we do not claim optimality of the induced grammar, and we perform no model
selection over a formally specified code family. The codec is \emph{exact},
not a lossy summary: its role is fidelity, not storage. In practice it does \emph{not} beat generic
gzip on size (Section~\ref{sec:experiments:codec}), which is itself a reported finding.

\subsection{Training arms}
\label{sec:method:arms}

Each seed trains three arms with identical hyperparameters and reset seeds:
\begin{itemize}
  \item \emph{Baseline:} unconstrained REINFORCE.
  \item \emph{Audit-only observer:} identical to the baseline (same weights, same trace hash) but the
        rule observer runs alongside and emits a rule bank from the training trace. Its actor is
        byte-identical to the baseline actor, which we verify explicitly.
  \item \emph{Constrained:} the action distribution is masked so that only actions admitted by rules
        extracted from the agent's own earlier trace can be selected during training.
\end{itemize}
The constrained arm was part of the original design intuition (\emph{let the rules steer the
training}) and is retained in the experiment to quantify the cost of that intuition
(Section~\ref{sec:evolution:constraint}).

\subsection{Rule-level policy fusion}
\label{sec:method:fusion}

Given rule banks $\{\mathcal{R}_i\}_{i=1}^{n}$ from $n$ seeds, the fused policy makes a decision at
state $s$ as follows (Figure~\ref{alg:fusion}). Let
$\mathcal{C}(s)=\{\,r\in\textstyle\bigcup_i \mathcal{R}_i : \mathbf{c}_r=\varphi(s)\,\}$ be the set of
candidate rules whose condition matches the current state. If $\mathcal{C}(s)=\varnothing$, the
state is a \emph{blind spot} and the fused policy falls back to the action of the actor-logit ensemble
(the argmax of the mean action logits over the $n$ source actors). Otherwise, the fused action is the action
of the candidate with the highest confidence; ties are broken by support, then by mean reward, then
by source index. Formally,
\begin{equation}
F(s) \;=\; \begin{cases}
\displaystyle \arg\max_{a}\,\max_{r\in\mathcal{C}(s)} w(r)\,\mathbf{1}[a_r=a], & \mathcal{C}(s)\neq\varnothing,\\[2mm]
\pi_{\mathrm{fb}}(s), & \mathcal{C}(s)=\varnothing,
\end{cases}
\label{eq:fusion}
\end{equation}
where $w(r)=(f_r,\ n_r,\ \mu_r,\ -i)$ in lexicographic order and $\pi_{\mathrm{fb}}$ is the blind-spot
fallback (the actor-logit ensemble argmax). Two per-decision quantities are recorded for every step:
\begin{equation}
\mathrm{cov}(s)=\mathbf{1}[\mathcal{C}(s)\neq\varnothing], \qquad
\mathrm{confl}(s)=\mathbf{1}\big[\#\{a_r : r\in\mathcal{C}(s)\}>1\big].
\label{eq:coverage}
\end{equation}
We call $\mathrm{confl}(s)=1$ a \emph{conflict}: the seeds disagree about the action for this
condition.

\paragraph{Decision classes.}
Every fused decision is classified into one of five classes, and the class is recorded alongside the
action: \emph{agreement} (all candidates agree), \emph{single-source} (only one bank contributes),
\emph{conflict} (candidates disagree on the action), \emph{blind spot} (no candidate; fallback used),
and \emph{fallback} (the residual single-actor action). These classes are not report formatting; they
are the conditions under which the result is interpreted. Reporting a fused return without the
coverage, conflict rate, blind-spot rate, fallback rate, source distribution, and arbitration outcome
is incomplete.
\begin{figure}[t]
\centering
\caption{Rule-level policy fusion with provenance record.}
\label{alg:fusion}
\begin{minipage}{0.9\textwidth}
\small
\textbf{Input:} rule banks $\mathcal{R}_1,\ldots,\mathcal{R}_n$; symbolizer $\varphi$; fallback actor $\pi_{\mathrm{fb}}$.\newline
For each decision at state $s$:
\begin{enumerate}
  \item $\mathbf{c}\leftarrow \varphi(s)$
  \item $\mathcal{C}\leftarrow\{\,r\in\bigcup_i\mathcal{R}_i : \mathbf{c}_r=\mathbf{c}\,\}$ \hfill (candidate rules)
  \item \textbf{if} $\mathcal{C}=\varnothing$ \textbf{then return} $\pi_{\mathrm{fb}}(s)$ and mark a \emph{blind spot}
  \item \textbf{if} all $r\in\mathcal{C}$ share one action \textbf{then} choose it and mark \emph{agreement} (or \emph{single-source} if $|\{i:r\in\mathcal{R}_i\}|=1$)
  \item \textbf{else} choose $r^{*}\leftarrow \arg\max_{r\in\mathcal{C}}\ (f_r,\ n_r,\ \mu_r,\ -i)$, attach its source bank and mark \emph{conflict}
  \item append $(c,a,\nu)$ to the decision log, update the chain digest, and return $(a,\nu)$
\end{enumerate}
\end{minipage}
\end{figure}

\subsection{Analysis instruments}
\label{sec:method:instruments}

Three instruments probe what the rules and the fused policy actually mean.

\emph{Behavioral agreement.} For two actors $\pi_i,\pi_j$, sample $N$ fresh states from each
actor's own occupancy and measure the proportion on which the two policies select the same action:
\begin{equation}
\bar{A}(\pi_i,\pi_j) \;=\; \frac{1}{N}\sum_{s\in\mathcal{S}_{ij}} \mathbf{1}[\pi_i(s)=\pi_j(s)],
\label{eq:agreement}
\end{equation}
where $\mathcal{S}_{ij}$ is the pooled fresh-state sample. Because the states are drawn by
occupancy rather than uniformly, we also report an episode-cluster-balanced estimate and a
cluster-balanced bootstrap interval.

\emph{Rule-set overlap.} Rule banks are compared with the Jaccard similarity of their exact
state--action sets,
\begin{equation}
J(\mathcal{R}_i,\mathcal{R}_j) \;=\; \frac{\lvert \mathcal{R}_i \cap \mathcal{R}_j \rvert}{\lvert \mathcal{R}_i \cup \mathcal{R}_j \rvert},
\label{eq:jaccard}
\end{equation}
together with the number of shared conditions whose recommended actions conflict.

\emph{Fitted-Q GPI baseline.} To place the fusion comparison on a principled footing, we fit, for
each frozen source actor $\pi_i$, an action-value regressor by fitted Q evaluation (FQE)
\cite{le2019batch,voloshin2021empirical} over a pooled replay buffer:
\begin{equation}
\hat{Q}^{(k+1)}(s,a) \;=\; r + \gamma\,\hat{Q}^{(k)}\big(s',\,\arg\max\nolimits_{a'}\pi_i(s')\big),
\label{eq:fqe}
\end{equation}
minimized in the least-squares sense over the replay transitions, with a reservoir cap of 25,000
transitions per seed and $\gamma=0.99$. The baseline policy is the generalized policy improvement
(GPI) selection \cite{sutton2018reinforcement}
\begin{equation}
\pi_{\mathrm{GPI}}(s) \;=\; \arg\max_{a}\,\max_{i}\,\hat{Q}_i(s,a).
\label{eq:gpi}
\end{equation}
We emphasize that this is an \emph{approximate} GPI: the classical GPI bound requires bounded value
estimation error, which we do not claim to control; the fitted critics are diagnostics, and their
failure is informative but not evidence about exact GPI.

\subsection{Matched-support causal symbol intervention}
\label{sec:method:causal}

The communication experiment (Section~\ref{sec:evolution:comm}) is kept methodologically separate from
rule fusion. In the redesigned \emph{private-target} game a sender observes a private label and a
receiver must act on the message alone; reward is given only for correct identification, so ignoring
the message cannot be optimal. We measure the causal effect of a learned symbol with a
\emph{matched-support intervention}: take states where the sender emits a particular message, replace
the message with a shuffled one (a permutation null), and measure the change in the receiver's action
probability. The intervention is matched so that only the message content changes. Under the controlled
intervention protocol, which holds the receiver state and task context fixed to the extent implemented
by the environment, we estimate the effect of replacing the discrete message on the receiver's action
probability. We distinguish observational association from intervention-based effects: in line with
causal approaches to explaining RL behavior \cite{madumal2020explainable}, the communication study
replaces the candidate symbol while holding the task state and receiver context fixed as defined by the
environment, which identifies only the local intervention effect in the designed game. It does not
establish that the learned symbol has a human-interpretable meaning. A staged acceptance gate governs
every causal claim: diversity $\rightarrow$ positive control $\rightarrow$ necessity $\rightarrow$
generalisation $\rightarrow$ causality $\rightarrow$ channel exclusivity, each with a pre-registered
verdict. The result is bounded to this designed task and depends on the stated common-support and
no-private-channel checks: it is \emph{not} evidence about state--action
rule fusion, and it does \emph{not} generalize to arbitrary neural policies.

\subsection{Statistical conventions}
\label{sec:method:stats}

All composition evaluations use 100 held-out episodes per policy, with reset seeds
$20{,}000{,}000 + i$ for $i=0,\ldots,99$; every policy sees the same episodes, so policy comparisons
are paired at the episode level. Paired differences are summarized by the mean of the per-episode
difference, with a 95\% percentile bootstrap interval over episodes (5{,}000 resamples). Standard
deviations are cross-episode sample standard deviations; we do not claim cross-seed variance
estimates, because the seed pool ($n=8$) is too small for that purpose. Where a quantity is
deterministic (replay equality, hash validity), we state that variance is structurally zero.

% ----------------------------------------------------------------------
\section{From Intuition to Protocol: a Failure-Driven Design History}
\label{sec:evolution}

This section documents, in order, the failures that shaped the protocol. Each entry states the
failure, its cause, and the change it forced. The sequence is part of the scientific record: it
explains why the protocol looks the way it does, and it is evidence about which intuitions are
unsafe.

\subsection{Compression, replay, and explanation are different claims}
\label{sec:evolution:replay}

The first intuition is that a recorded trace \emph{is} the explanation: ``here is exactly what the
agent did.'' A trace is faithful, but it is not a description. A full trace of a single seed on
CartPole-v1 contains $\approx$97{,}646 transitions (and up to 300{,}000 records per arm across the
population); a human cannot inspect it, two traces cannot be compared except by exact matching, and
a trace says nothing about states the agent never visited. The lesson, formalized as
Section~\ref{sec:method:ledger}, is that traces belong to the fidelity layer, and a separate
compression step is needed for the description layer. We therefore treat \emph{trace} and
\emph{rule bank} as distinct artifacts with distinct evidentiary roles.

\subsection{First-generation prototype failures}
\label{sec:evolution:v1}

An initial prototype attempted to compress behavior with sequence grammar induction. Three defects
were found, each independently fatal for the intended claim:

\begin{enumerate}
  \item \emph{Lossy grammar.} The prototype's sequence-compression inducer was a simplified variant
        of SEQUITUR \cite{nevillmanning1997identifying} and was not equivalent to the published
        algorithm. The induced grammar could not be proven to reproduce the source sequence, so any
        rule extracted from it was not anchored to the trace. Cause: reimplementing a linear-time
        grammar algorithm from memory without a conformance test. Fix: discard grammar-based
        compression as an evidence path; compress only \emph{statistics} (support, confidence, mean
        reward) whose computation is a deterministic reduction of the trace.
  \item \emph{Conflation of one-step and window rules.} The observer initially treated ``the agent
        chose $a$ in condition $\mathbf{c}$'' and ``the agent repeated a pattern across several
        steps'' as the same kind of rule. They are different quantities: the first is a per-decision
        statistic, the second a sequence property. Mixing them produced rules with no well-defined
        estimand. Cause: one object was used for two jobs. Fix: the protocol admits only per-decision
        condition--action rules (\ref{eq:rulestats}); sequence structure is outside the evidence
        budget of this paper.
  \item \emph{``Replay'' meant different things.} An early verifier checked that recorded actions,
        when replayed, reproduced \emph{some} transition statistics; it did not check exact
        per-transition equality. Cause: the verifier tested aggregate health, not fidelity. Fix:
        exact float32 equality per transition (\ref{eq:replay}), replayed from the recorded reset
        seed.
\end{enumerate}

\subsection{Rule-constrained training interferes with behavior}
\label{sec:evolution:constraint}

The original design intuition was that induced rules should actively guide training: mask out
actions that violate the agent's own rules. The implemented mask is state-conditioned and hard. On a
state symbol that a rule covers, every other action is set to $-\infty$, so the sampled action is the
rule's action with probability one; on an uncovered symbol the action set is untouched. This
supersedes an earlier \emph{global} action-pruning variant that could remove an action permanently
merely because it was absent from frequent rules.

A two-seed pilot ($\tau_c=0.90$) was the first negative result, and it was sign-inconsistent rather
than uniformly catastrophic: on seed 11 the constrained actor's held-out return fell from $500$ to
$121.3$ ($-378.7$), while on seed 29 it \emph{rose} from $131.0$ to $162.5$ ($+31.5$). The induced
bank covered only $0.41\%$ of held-out steps on seed 11 and $6.98\%$ on seed 29, so the intervention
was close to inert on both. A high-confidence threshold does not make a constraint safe; it makes it
rarely applied.

\begin{figure}[t]
\centering
\includegraphics[width=0.78\textwidth]{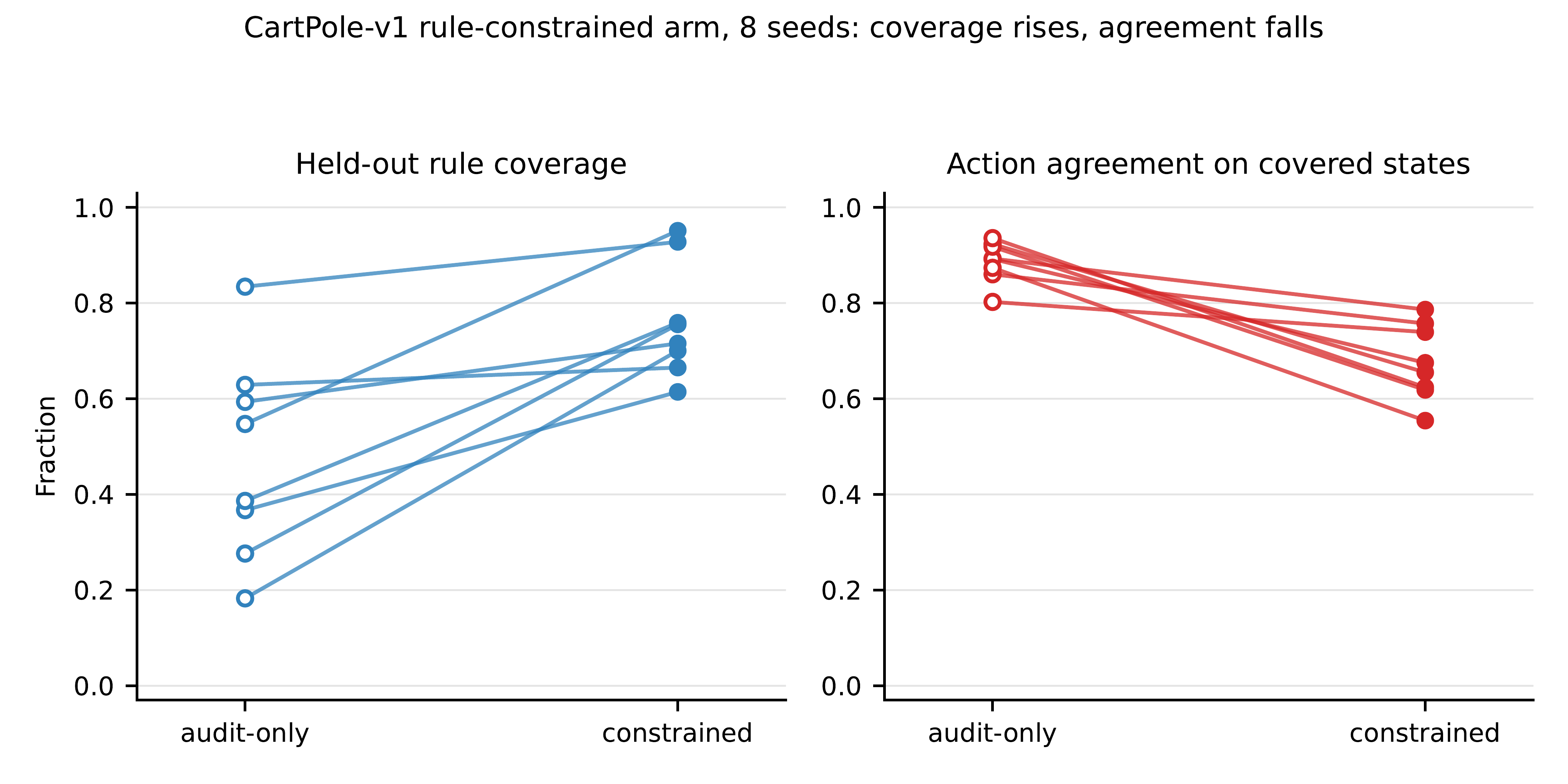}
\caption{CartPole-v1 rule-constrained arm, eight seeds, $\tau_c=0.70$. Left: held-out coverage of the
induced rule bank. Right: action agreement between the bank and the learned policy on covered states.
Each line is one seed; hollow markers are the audit-only arm and filled markers the constrained arm.
Every arm is measured on its own state occupancy, so this is a same-seed pairing, not a matched-state
comparison. Coverage rises in $8/8$ seeds and agreement falls in $8/8$ seeds: the mask does not reduce
the reach of the rule bank, it removes the policy's ability to disagree with it.}
\label{fig:constraint-coverage}
\end{figure}

The eight-seed study ($\tau_c=0.70$) confirms the heterogeneity
(Table~\ref{tab:perseed}, Figure~\ref{fig:constraint-coverage}): on CartPole-v1, constraining training
improved some seeds ($29$: $131.0\rightarrow293.5$; $71$: $130.2\rightarrow387.2$; $43$:
$153.75\rightarrow500.0$) and damaged others ($11$: $500.0\rightarrow182.3$; $149$:
$500.0\rightarrow96.1$; $307$: $500.0\rightarrow161.2$). On Acrobot-v1 every seed's held-out return is
$-500$ in both arms; that value is the step limit, not a measured difference, so the task contributes
a null result rather than evidence of damage.

\emph{The cause is structural, and it is not a coverage failure.} The threshold-relaxation experiment
was designed to test the hypothesis that low coverage caused the damage, and it refutes that
hypothesis: across the eight seeds the induced bank's held-out coverage \emph{rises} in $8/8$ seeds
(mean $0.477\rightarrow0.761$) while action agreement on covered states \emph{falls} in $8/8$ seeds
(mean $0.887\rightarrow0.676$; Figure~\ref{fig:constraint-coverage}). What contracts is the policy's
own state occupancy, not the reach of the bank. On the training trace, where the learner does its
exploring, the effective number of state symbols visited, $2^{H}$, falls in $7$ of $8$ seeds (seed
$11$: $79.3\rightarrow58.4$; $149$: $91.1\rightarrow45.4$; $307$: $70.5\rightarrow37.5$; $43$:
$83.8\rightarrow63.5$; the exception is seed $211$, $53.1\rightarrow60.4$). The same statistic is
mixed on held-out evaluation, where it contracts in only $4$ of $8$ seeds: a policy that fails early
visits more of the state space before it dies, so held-out occupancy conflates where the policy
lives with how long it survives.

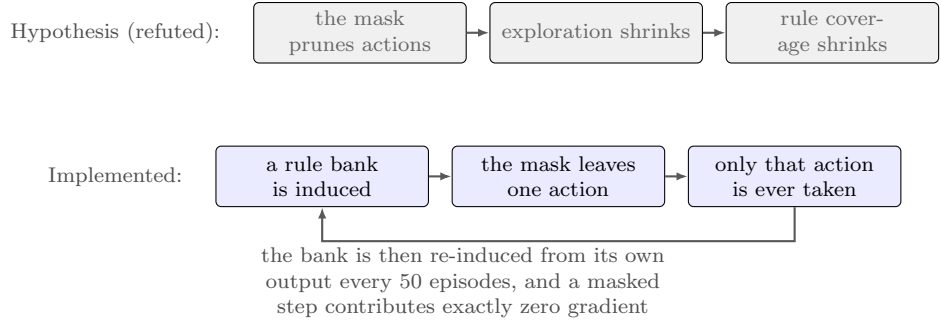
\begin{figure}[t]
\centering
\begin{tikzpicture}[
  stage/.style={draw, rounded corners=2pt, align=center, font=\scriptsize,
                text width=26mm, minimum height=8mm, inner sep=3pt},
  refuted/.style={stage, fill=black!6, text=black!60},
  actual/.style={stage, fill=blue!8},
  flow/.style={-{Latex[length=1.5mm]}, draw=black!65, thick}
]
\node[font=\scriptsize, text=black!70] (hlab) at (0,0) {Hypothesis (refuted):};
\node[refuted, right=3mm of hlab] (h1) {the mask prunes actions};
\node[refuted, right=3mm of h1] (h2) {exploration shrinks};
\node[refuted, right=3mm of h2] (h3) {rule coverage shrinks};
\draw[flow] (h1) -- (h2);
\draw[flow] (h2) -- (h3);
\node[font=\scriptsize, text=black!70] (llab) at (0,-1.9) {Implemented:};
\node[actual, right=3mm of llab] (l1) {a rule bank is induced};
\node[actual, right=3mm of l1] (l2) {the mask leaves one action};
\node[actual, right=3mm of l2] (l3) {only that action is ever taken};
\draw[flow] (l1) -- (l2);
\draw[flow] (l2) -- (l3);
\draw[flow] (l3.south) -- ++(0,-0.45) -| (l1.south);
\node[font=\scriptsize, text=black!70, align=center, text width=76mm] (note)
  at (4.6,-3.3) {the bank is then re-induced from its own output every $50$ episodes, and a masked
  step contributes exactly zero gradient};
\end{tikzpicture}
\caption{Mechanism of the training-time constraint, as implemented. The top row is the hypothesis
that the threshold-relaxation experiment refutes. The bottom row is the loop the code implements: a
hard one-hot mask, whose log-probability at a masked step is exactly zero and which therefore
contributes nothing to the policy gradient, together with a rule bank re-induced from the masked
trajectory every $50$ episodes, which makes the mask's own output the evidence for its own rules.}
\label{fig:constraint-mechanism}
\end{figure}

Two mechanisms are built into the implementation and together remove the learning signal the
constraint is meant to shape (Figure~\ref{fig:constraint-mechanism}). We establish that both are
present; we do not apportion the damage between them. First, the mask deletes its own learning
signal: when one action is left at probability one its log-probability is exactly zero, so a masked
step contributes exactly zero to the REINFORCE objective $-\log\pi(a\mid s)\,G$. Between $36.7\%$ and
$74.4\%$ of training steps were masked in our runs, so up to three quarters of the gradient budget
was discarded at precisely the states the policy visited most often. This half is arithmetic rather
than empirical: no experiment can strengthen it. Second, the observer re-enters the loop it is meant
to observe. The rule bank is re-induced from the constrained trajectory every $50$ episodes, and that
trajectory contains only the actions the mask permitted, so the mask's own choices become the
evidence that re-admits the same rules. Measured directly, $68\%$ of the observations the final bank
rests on were generated by the mask itself (range $45\%$--$85\%$ across seeds). The observer's counts
are cumulative and are never reset, so a covered rule cannot be falsified by any later evidence:
across all eight seeds and six re-induction generations we observe $0$ retractions of an admitted
rule and $0$ changes to a rule's forced action. The loop is self-confirming in the precise sense that
the data which would falsify a rule are never collected.

\emph{Two measurements, two different questions.} The rule-constrained arm is evaluated with the rule
action substituted whenever a rule fires, which is $61\%$--$95\%$ of held-out steps. The return it
reports is therefore that of the \emph{hybrid} system---the learned policy and the frozen bank
together, running as the implementation runs them---not of the network alone. We report both
quantities, because they answer different questions and neither supersedes the other: the hybrid
score asks what this intervention delivers when the bank is allowed to speak, and the network's own
score asks what the learner retained. To obtain the second we re-evaluated the same saved checkpoints
on the same held-out reset seeds with the substitution disabled, following the protocol prior work
used to study the removal of a trained mask \cite{huang2022masking}. The two readings disagree in an
informative way (Table~\ref{tab:attribution}). As a hybrid the intervention costs $55.45$ on average
($271.6$ against $327.0$ for the unconstrained actors); stripped of the bank, the network is further
below those actors ($225.13$), and four seeds lose more than $120$ points. The difference between the
two readings is the bank's own contribution, which is positive on average ($+46.44$) and decisive on
seed $149$, where a network that alone scores $486.4$ is scored $96.1$ once the bank overrides it.
Quoting only the hybrid number would understate the cost to the learned policy; quoting only the
network's own number would imply the bank is uniformly harmful, which it is not.

The same rule machinery that \emph{composes} well offline (Section~\ref{sec:experiments:cartpole}) is
not dependable as a training-time intervention. Consequence: the protocol separates rule induction
(audit-only observer) from rule composition (offline fusion); the constrained arm is retained as a
quantified negative result that is heterogeneous and sign-inconsistent across seeds, not as evidence
that induced rules are generally harmful.

\begin{table}[t]
\centering
\caption{Per-seed held-out returns on CartPole-v1 (20 held-out episodes per seed, reset seeds
$10^{4}\!\cdot\!\text{seed}+9\!\times\!10^{5}+i$; each seed is evaluated on its own reset-seed range,
so these rows are not the matched 100-episode comparison set of Table~\ref{tab:primary}): baseline
actor versus
the rule-constrained actor. The constraint is a training-time intervention; its effect is
heterogeneous and is negative for every seed that reaches the ceiling without it. On Acrobot-v1
every seed returns $-500$ in both arms.}
\label{tab:perseed}
\begin{tabular}{lccc}
\toprule
Seed & Baseline & Constrained & Change \\
\midrule
11 & 500.0 & 182.3 & $-317.7$ \\
29 & 131.0 & 293.5 & $+162.5$ \\
43 & 153.75 & 500.0 & $+346.25$ \\
71 & 130.15 & 387.15 & $+257.0$ \\
101 & 201.3 & 149.15 & $-52.15$ \\
149 & 500.0 & 96.1 & $-403.9$ \\
211 & 500.0 & 403.2 & $-96.8$ \\
307 & 500.0 & 161.2 & $-338.8$ \\
\bottomrule
\end{tabular}
\end{table}

\begin{table}[t]
\centering
\caption{Two held-out readings of the rule-constrained arm on CartPole-v1, one per column group.
``Hybrid'' is what the arm reports: the composed system, bank included. ``Network only''
re-evaluates each constrained arm's saved checkpoint with the rule substitution disabled, so it is
the learned policy's own return. ``Unconstrained'' is the baseline actor on the same reset seeds. The
two readings answer different questions and are not alternatives to one another. A positive
substitution gain means the frozen rule bank scored better than the network it was induced from.}
\label{tab:attribution}
\begin{tabular}{lcccc}
\toprule
Seed & Hybrid & Network only & Unconstrained & Substitution gain \\
\midrule
11 & 182.3 & 127.8 & 500.0 & $+54.5$ \\
29 & 293.5 & 365.1 & 131.0 & $-71.6$ \\
43 & 500.0 & 217.9 & 153.8 & $+282.1$ \\
71 & 387.1 & 237.6 & 130.2 & $+149.6$ \\
101 & 149.2 & 80.0 & 201.3 & $+69.2$ \\
149 & 96.1 & 486.4 & 500.0 & $-390.2$ \\
211 & 403.2 & 120.8 & 500.0 & $+282.4$ \\
307 & 161.2 & 165.4 & 500.0 & $-4.2$ \\
\midrule
Mean & 271.6 & 225.1 & 327.0 & $+46.4$ \\
\bottomrule
\end{tabular}
\end{table}

\subsection{Confidence thresholds and the 100\% blind-spot collapse}
\label{sec:evolution:threshold}

Rule admission requires a confidence threshold. A high-confidence variant ($\tau_c=0.90$) produced
fusion policies that deferred to the fallback actor on 100\% of held-out decisions: the strict
threshold admitted so few rules that no condition was covered, and the ``fusion'' was the fallback
in disguise. Cause: confidence is a conditional statistic; raising it shrinks support monotonically,
and at the extreme the rule bank no longer covers the state space. The threshold was therefore set
to $\tau_c=0.70$ with $\tau_s=8$ \emph{and} an explicit blind-spot fallback (\ref{eq:fusion}) so
that coverage failure is observable rather than silent. The episode is a general lesson:
\emph{coverage is not correctness}. Confidence, support, coverage, action agreement, and return are
five different quantities; a high coverage or confidence threshold is a statistical choice, never a
semantic validation of the rule. This is a two-setting exploratory choice,
not a systematic threshold search; we report the boundary in
Section~\ref{sec:experiments}.

\subsection{Serialization and ledger-schema failures}
\label{sec:evolution:serialization}

Two failures concerned the audit artifacts themselves. First, a human-readable (pretty-printed)
serialization of the trace inflated its size by roughly $1.8\times$; after switching to a compact
canonical encoding, the compressed trace ledger was still 8.4\% larger (CartPole-v1) and 9.9\%
larger (Acrobot-v1) than the gzip-compressed raw trace, because the grammar-style codec adds
structure without removing bytes. The codec is nevertheless exact: the decoded payload hashes to the
input hash. We report these costs as part of practical reproducibility. Second, the update-ledger
schema initially allowed a sidecar field (a grammar-generation annotation) to vary without changing
the optimizer record, so two runs with identical optimizer behavior could hash differently.
Cause: the hash covered non-core fields. Fix: the ledger hashes only a normalized core record
(step index, parameter gradient hash, pre/post checkpoint hashes), making equivalent runs hash
identically.

The same lesson appears in the communication task below (Section~\ref{sec:evolution:comm}): a sidecar
field (a grammar-generation annotation) once entered the update hash even though it did not affect
optimisation, and the fix was to hash only a normalized core record. An audit instrument must audit
itself.

\subsection{Arbitration sensitivity and the random-selection anomaly}
\label{sec:evolution:arbiter}

The confidence-ranked arbiter was chosen as the default resolver because it is the natural
generalization of the per-rule confidence statistic. The results (Section~\ref{sec:experiments:acrobot})
show it is the wrong default on Acrobot-v1: uniformly random arbitration among candidate rules beats
every deterministic variant tested, including the deployed one. We report this as an \emph{open
mechanism} rather than a method. A1b later resolves the anomaly as an induction/deployment protocol
mismatch: the bank was induced from sampled training actions but deployed under argmax. Re-inducing from
argmax-consistent (greedy) labels reverses the ordering (confidence-ranked fusion $-141.77$ beats random
$-234.26$), so ``random beats confidence'' is a property of the sampled-bank protocol, not of confidence
arbitration in general. Treated as a design event, it demotes the arbiter from a trusted
component to a diagnostic: the fused policy is reported together with its full decision mix, and the
random-selection ablation is part of the result, not a footnote.

\subsection{Communication-task redesign}
\label{sec:evolution:comm}

The causal-symbol evidence rests on a designed coordination game that had to be rebuilt twice. The
original \emph{parity} game (sender sees a digit, receiver acts on the message, reward on parity match)
collapsed to a degenerate equilibrium: both policies settled on the constant action C/C, the unique
joint-reward optimum under both parities, so the channel carried nothing. Under a parity-matching
reward, constant actions are a Nash equilibrium and no optimizer can escape an objective whose global
optimum ignores the channel---the failure was in \emph{task design}, not learning.

A second, abandoned early design (the private-target variant) failed for a different reason: the
source review found the original coordination payoff made the same joint action optimal for both
parities; a blind control masked only the label while the image encoder retained parity information;
a reserved index lay outside the embedding range; the codebook was bypassed by a straight-through
continuous path; and an opponent-prediction branch was degenerately solvable by copying. The fix was a
parity-critical payoff with opposed optima, joint masking of label and image pathways, corrected index
handling, straight-through codebook wiring with perplexity reporting, and the staged gate above.

Even the redesigned game initially appeared to fail: after training, receiver accuracy was $0.500$
(chance) and the estimated causal influence $\Delta P(C)=0.00088$ ($95\%$ CI $[-0.00136,0.00320]$) was
indistinguishable from zero. The cause was \emph{optimization exposure}, not architecture: the run had
performed only $79$ optimizer updates; reducing the batch size from $128$ to $8$ raised the count to
$1{,}250$, and a run with $12{,}500$ updates succeeded decisively. Two analysis methods also produced
impressive but empty numbers: a ``symbol lookup'' analysis was tautological (the action decoder read
the first token directly), and a whole-sequence mutual-information estimate saturated at $1$ bit
because sparse, high-cardinality messages make plug-in MI vacuous. Both were replaced by a per-position
MI against a permutation null. The only surviving causal claim comes from the redesigned task and is
bounded to it.

% ----------------------------------------------------------------------
\section{Experiments and Results}
\label{sec:experiments}

\subsection{Setup}
\label{sec:experiments:setup}

Experiments use CartPole-v1 (4 state dimensions, 2 actions) and Acrobot-v1 (6 dimensions, 3
actions) from the Gymnasium suite \cite{towers2024gymnasium}. Eight seeds
$\{11,29,43,71,101,149,211,307\}$ are used; each seed trains 300 episodes per arm with REINFORCE
(\ref{eq:reinforce}) and a shared frozen symbolizer (\ref{eq:symbolizer}) calibrated from 24
random-policy episodes per seed. Rules are admitted with $(\tau_s,\tau_c)=(8,0.70)$. Composition
evaluates 100 held-out episodes per policy on the common reset seeds. All reported traces were
post-hoc audited: hash chains, update ledgers, baseline/observer equivalence, and environment
replay were re-verified after the runs completed.

Because deep RL results can vary substantially with random seeds, environment nondeterminism,
evaluation protocol, and reporting choices \cite{henderson2018deep}, we report seed identities,
matched reset seeds, evaluation distributions, and uncertainty intervals separately rather than
folding them into a single number: the seed pool is small ($n=8$), the reported intervals are over
episodes and not over seeds, and the comparator for the headline comparison is selected post hoc
(Section~\ref{sec:discussion:limitations}).

\subsection{Registered predictions}
\label{sec:experiments:pred}

We fix predictions before reporting any result, so that they can be contradicted. They are
\emph{unregistered} (no external preregistration record exists), so they are treated as post hoc when
assessing claim strength. Each receives a status in Section~\ref{sec:experiments:verdicts}.

\begin{table}[t]
\centering\small
\caption{Predictions fixed before analysis. Layer: operational-health (OH), diagnostic (D),
identification-bearing (ID). Status assigned in Section~\ref{sec:experiments:verdicts}.}
\label{tab:pred}
\begin{tabular}{@{}l p{0.44\linewidth} l l@{}}
\toprule
ID & Prediction & Layer & Status \\
\midrule
P1 & Hash-chain, ledger and replay checks pass on all recorded runs. & OH & supported \\
P2 & The lossless codec reconstructs every trace prefix exactly. & OH & supported \\
P3 & Baseline and audit-only arms are training-equivalent. & OH & supported \\
P4 & Cross-seed rule overlap is substantially above zero under a shared symbolizer. & D & supported \\
P5 & Two actors with high rule overlap also agree on most fresh held-out states. & ID & contradicted \\
P6 & Fused mean return exceeds the best single source actor on both tasks. & ID & comparator-dependent$^{\ddagger}$ \\
P7 & Blind spots, not conflicts, are the main obstacle to fusion. & ID & contradicted \\
P8 & The deployed confidence-ranked arbiter is at least as good as alternative selectors. & ID & induction-protocol-dependent$^{\S}$ \\
P9 & A learned rule bank predicts an independent holdout above an action-frequency baseline, with
coverage, agreement, and confidence reported separately. & ID & contradicted (at floor) \\
\bottomrule
\end{tabular}
\end{table}

$^{\ddagger}$P6 holds on CartPole-v1 only under the narrow reading of ``best single actor'' as the
actor selected by training return. Against the best \emph{held-out} actor on the same matched episodes
it fails by $294.04$ return points (Table~\ref{tab:comparators}), so we record the prediction as
comparator-dependent rather than supported. On Acrobot-v1, under \emph{argmax} deployment, all eight
source actors tie at the floor, so the two readings coincide there and the prediction holds (under the
sampled evaluation protocol the best single actor and the ensemble do not tie; see
Table~\ref{tab:protocol}).

$^{\S}$P8 is contradicted under the original sampled-bank induction (confidence-ranked fusion $-443.37$
vs.\ random $-187.02$) but \emph{supported} under protocol-consistent greedy-bank induction ($-141.77$
vs.\ random $-234.26$); the verdict therefore depends on the induction protocol, not on the arbiter
alone (A1b).

P9 (the action-frequency baseline prediction) is reported here rather than deferred, because
coverage/agreement statistics are otherwise undefined relative to a floor: a rule bank that agrees
with the actor on $51\%$ of states is only informative once compared with the majority-action
frequency of the same states. On the same 10{,}000 fresh states used for the agreement study
(Section~\ref{sec:experiments:agreement}), the occupancy-weighted majority-action baseline is
$51.32\%$. This baseline is a \emph{per-occupancy majority-action predictor}: on each actor's own
5000 fresh states we predict that actor's modal action---action 1 on the seed-11 occupancy
($2528/5000$) and action 0 on the seed-29 occupancy ($2604/5000$)---so the pooled correct rate is
$(2528+2604)/10000 = 51.32\%$, against an observed cross-actor agreement of $51.38\%$. The two
figures are computed on the same 10{,}000 states. The agreement therefore sits
essentially at the action-frequency floor (margin $+0.06$ points), so the rule bank does \emph{not}
predict the holdout above the baseline; the prediction is reported as contradicted.

\subsection{Audit integrity}
\label{sec:experiments:integrity}

Table~\ref{tab:integrity} reports the fidelity layer. Every hash chain is valid, every update
ledger verifies, baseline and audit-only actors are byte-identical, and every recorded transition
replays exactly under its reset seed. The lossless codec round-trips exactly in both environments.
These are operational-health results: they certify the artifacts, and they do not by themselves
support any claim about policy quality.

\begin{table}[t]
\centering
\caption{Audit integrity (operational-health layer). Values are deterministic; variance is
structurally zero.}
\label{tab:integrity}
\begin{tabular}{lcc}
\toprule
 & CartPole-v1 & Acrobot-v1 \\
\midrule
Trace records (all three arms) & 1{,}725{,}655 & 2{,}544{,}513 \\
Optimizer update records & 7{,}200 & 7{,}200 \\
Environment replay checks (post-hoc) & 1{,}122{,}403 & 1{,}742{,}570 \\
Constraint rule-reference verifications & 290{,}052 & 697{,}817 \\
Hash chains valid & yes & yes \\
Update ledgers valid & yes & yes \\
Baseline vs.\ audit-only actor identical & yes & yes \\
Codec round-trip exact & yes & yes \\
Raw trace bytes / gzip & 61.3\,MB / 12.3\,MB & 57.6\,MB / 12.5\,MB \\
Codec archive (gzip) & 13.4\,MB & 13.7\,MB \\
\bottomrule
\end{tabular}
\end{table}

\paragraph{Tamper probe.}
Hash validity is a claim about tamper detection, so we test it directly. We modified the reward field
in the first record of a \emph{copy} of the audited CartPole merged trace---the 100{,}000-decision,
$\tau_c=0.70$ fused trace whose 68{,}409 chosen rule references resolve to source grammars and whose
41{,}728 fallback references resolve to the selected seed-11 actor---and deliberately did \emph{not}
recompute the chain. The replay-and-provenance validator (Def.~\ref{def:audit}) rejected the copy at
step~1 with a hash mismatch, and the original trace was left unmodified. This is a tamper probe on the
audited CartPole trace only; it is not evidence that every fused trace in every environment has been
subjected to the same test.

\subsection{Rule structure across seeds}
\label{sec:experiments:structure}

Rule banks are seed-specific but structurally related (Table~\ref{tab:structure}). On CartPole-v1,
all eight seeds share a core of 38 conditions on which they \emph{agree} (same action in every
seed), and the mean pairwise Jaccard overlap of rule sets is 0.56; the two-seed comparison used
throughout the report is 0.58. On Acrobot-v1 the picture inverts: the all-seed intersection contains
only 2 conditions, and both are \emph{conflicting}; the mean pairwise Jaccard is 0.096. Across the
two environments, 77 of 98 shared conditions conflict in Acrobot-v1 versus 0 of 77 in CartPole-v1.

\begin{table}[t]
\centering
\caption{Rule structure across seeds. Pairwise statistics use all 28 seed pairs.}
\label{tab:structure}
\begin{tabular}{lcc}
\toprule
 & CartPole-v1 & Acrobot-v1 \\
\midrule
Rule count per seed (range) & 100--145 & 77--303 \\
Conditions shared by $\ge$2 seeds & 165 & 445 \\
All-seed intersection (conditions) & 38 & 2 \\
\quad same-action core & 38 & 0 \\
\quad conflicting & 0 & 2 \\
Mean pairwise rule Jaccard (\ref{eq:jaccard}) & 0.557 & 0.096 \\
Shared conditions (seeds 11 vs.\ 29) & 77 & 98 \\
\quad conflicting (seeds 11 vs.\ 29) & 0 & 77 \\
\bottomrule
\end{tabular}
\end{table}

\subsection{Composition on CartPole-v1}
\label{sec:experiments:cartpole}

\begin{figure}[t]
\centering
\includegraphics[width=0.72\textwidth]{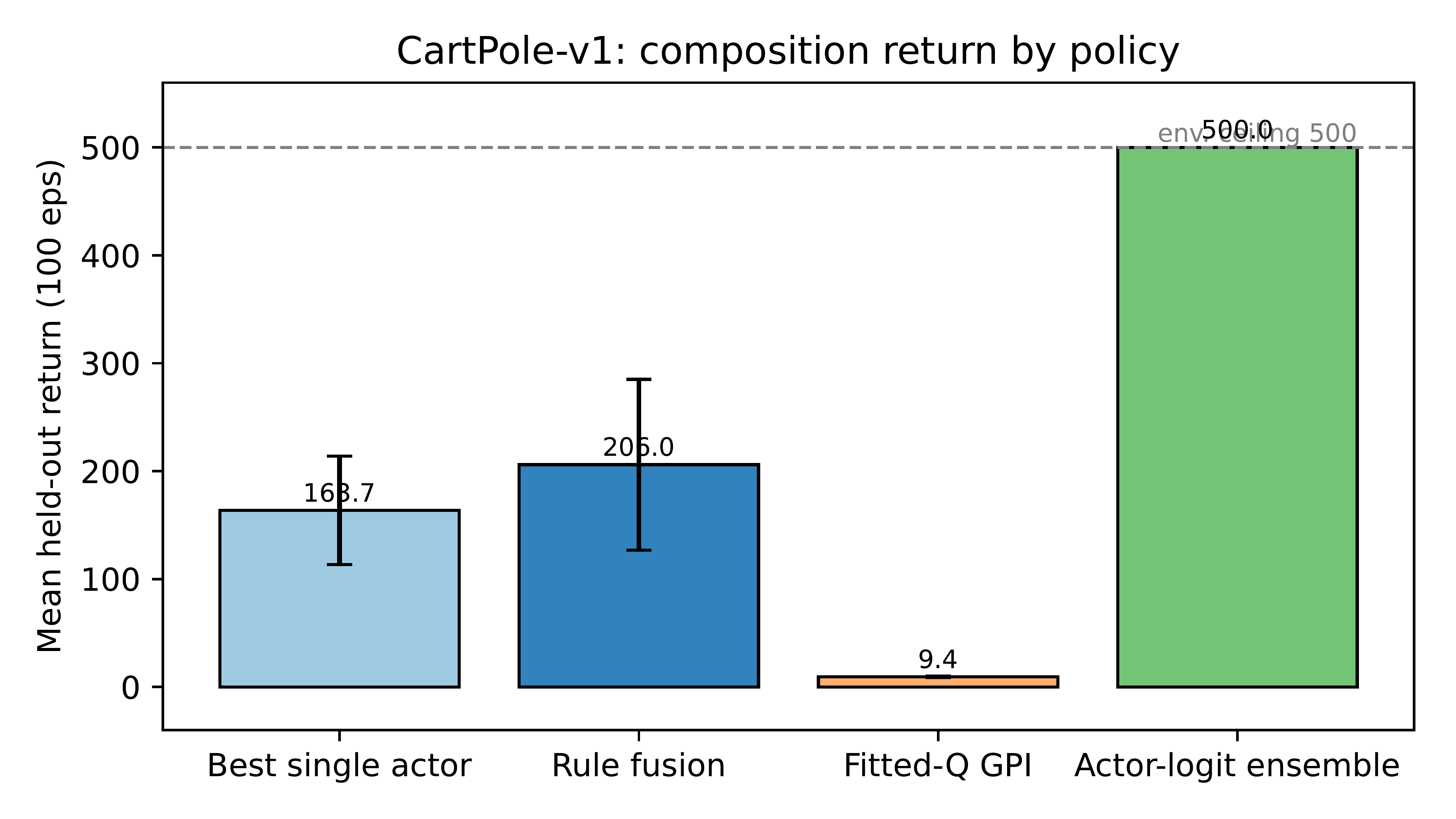}
\caption{CartPole-v1 composition comparison. Error bars are 95\% bootstrap intervals of the mean
over the 100 held-out episodes. The dashed line marks the environment ceiling (500). The
actor-logit ensemble and the FQE-GPI baseline bracket the fusion result from above and below;
FQE-GPI is reported here and analyzed as a failed diagnostic in
Section~\ref{sec:experiments:fqe}.}
\label{fig:cartpole}
\end{figure}

Table~\ref{tab:primary} and Figure~\ref{fig:cartpole} report the main composition result. On
CartPole-v1, the rule-level fusion policy reaches a mean held-out return of 205.96 (median 181,
cross-episode std 79.1) versus 163.65 for the single-actor comparator, a paired difference of $+42.31$
with 95\% bootstrap interval $[23.98,61.93]$. The comparator is not a pre-declared baseline: it is the
source actor with the highest \emph{training} return mean (seed 43), which is a post-hoc choice, and
the task exhibits ceiling effects---four of the eight actors reach the 500-step ceiling on their own
held-out evaluations. We therefore report this comparison as \emph{exploratory} rather than as a
pre-declared win over the best baseline, and Table~\ref{tab:comparators} reports how sensitive the
headline is to the selection rule. That sensitivity is large and runs against us: on the same matched
episodes, four of the eight individual actors reach the ceiling (500.0), the mean and median over the
eight actors are 330.86 and 358.81, and fusion (205.96) is $294.04$ points below the best held-out
actor ($-294.04$, 95\% interval $[-308.97,-278.16]$). Fusion exceeds only the comparator that the
training-return rule happens to select, and the gain is therefore a statement about that selection
rule, not about fusion being a strong single policy in this environment. Fusion decisions are
98.3\% rule-covered (20{,}240 of
20{,}596 steps) with only 75 conflicting decisions (0.4\%) and 356 blind spots (1.7\%), which is
consistent with the low-conflict rule structure of Table~\ref{tab:structure}. The actor-logit
ensemble saturates the environment ceiling (500.0, zero variance), which makes it a ceiling-level
comparison rather than a differentiating one.

\begin{table}[t]
\centering
\caption{Primary composition results. Means and paired differences over 100 common held-out
episodes; 95\% bootstrap intervals in brackets. ``Best single actor'' is the source actor with the
highest training return mean (seed 43 on CartPole-v1, seed 211 on Acrobot-v1); the selection is post
hoc and is analyzed in Table~\ref{tab:comparators}.}
\label{tab:primary}
\begin{tabular}{lcc}
\toprule
Policy & CartPole-v1 & Acrobot-v1 \\
\midrule
Best single actor & 163.65 & $-500.0$ \\
Actor-logit ensemble & 500.0 (ceiling) & $-500.0$ \\
Grammar fusion $F$ (\ref{eq:fusion}) & \textbf{205.96} & $-443.37$ \\
Fitted-Q GPI (\ref{eq:gpi}) & 9.34 & $-499.57$ \\
\midrule
Paired $\Delta$ fusion vs.\ best single & $+42.31\ [23.98,\,61.93]$ & $+56.63\ [39.32,\,75.15]$ \\
Paired $\Delta$ ensemble vs.\ best single & $+336.35$ & $0.0$ \\
Fusion coverage / conflicts / blind spots & 98.3\% / 75 / 356 & 95.7\% / 39{,}602 / 1{,}909 \\
\bottomrule
\end{tabular}
\end{table}

\begin{table}[t]
\centering
\caption{CartPole-v1 comparator sensitivity on the \emph{matched} evaluation set (the same 100 common
reset seeds used in Table~\ref{tab:primary}). Every row is evaluated on the identical reset seeds, so
the rows are directly comparable; the two single-actor rows differ only in the selection rule. The
seed-43 row and the ensemble row reproduce the recorded values of Table~\ref{tab:primary} exactly,
which validates the re-evaluation harness. Paired $\Delta$ is reported relative to rule fusion $F$
(Eq.~\ref{eq:fusion}); it is undefined for the population summaries.}
\label{tab:comparators}
\begin{tabular}{lcc}
\toprule
Comparator & Mean return & Paired $\Delta$ vs.\ fusion \\
\midrule
Best \emph{training}-return actor (seed 43) & 163.65 & $+42.31\ [23.98,\,61.93]$ \\
Best \emph{held-out} actor (seeds 11/149/211/307, tie) & 500.0 & $-294.04\ [-308.97,\,-278.16]$ \\
Mean over the eight actors & 330.86 & --- \\
Median over the eight actors & 358.81 & --- \\
Worst actor (seed 29) & 131.11 & $+74.85$ \\
Actor-logit ensemble & 500.0 & $-294.04$ \\
Rule fusion $F$ (\ref{eq:fusion}) & 205.96 & --- \\
\bottomrule
\end{tabular}
\end{table}

\subsection{Acrobot-v1: the conflict regime and the failure of confidence arbitration}
\label{sec:experiments:acrobot}

On Acrobot-v1, every source actor is at the floor ($-500$), yet the fusion policy is not: it
averages $-443.37$ (median $-500$, std 90.8), with 39 of 100 episodes terminating before the
500-step cap and a maximum of $-126$. The paired difference versus the best single actor is $+56.63$
$[39.32,75.15]$. However, the fusion decisions are 95.7\% rule-covered and 89.2\% \emph{conflicting}
(39{,}602 of 44{,}376 steps), so the fused policy is almost entirely an arbitration among
disagreeing seeds.

\begin{figure}[t]
\centering
\includegraphics[width=0.72\textwidth]{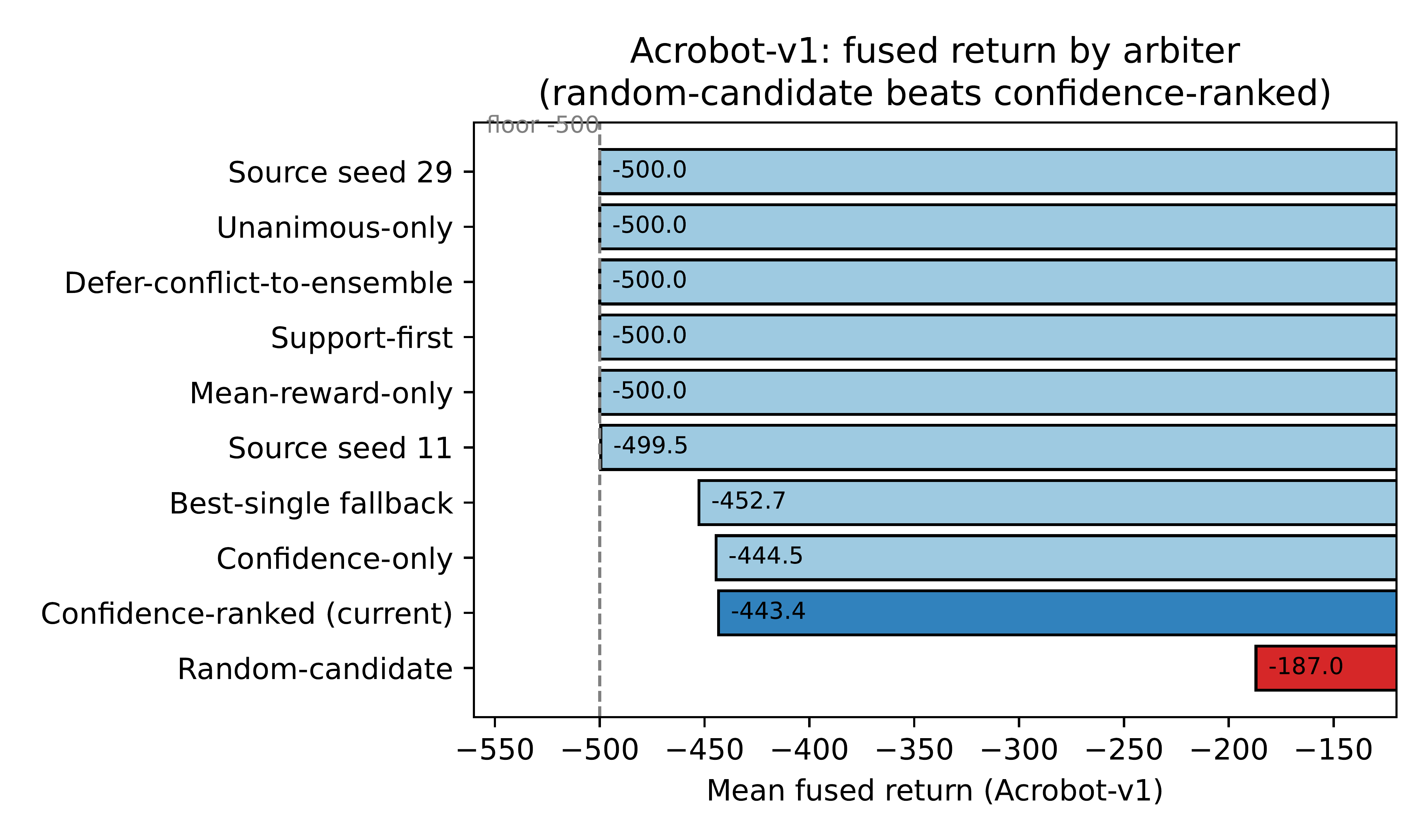}
\caption{Acrobot-v1 arbitration ablations on matched episodes. Random arbitration among candidate
rules (solid red bar) substantially outperforms confidence-ranked arbitration; the dashed line marks
the five-stream random-arbitration mean. Uniform random actions (ignoring the rules) average
$-498.71$ and are omitted for scale.}
\label{fig:acrobot}
\end{figure}

The arbitration rule matters enormously, and not in the intended direction
(Figure~\ref{fig:acrobot}, Table~\ref{tab:ablations}). Uniformly random arbitration among candidate
rules returns $-187.02$ on the primary evaluation (95\% paired interval $[231.38,278.21]$ relative
to confidence-ranked fusion) and is replicated across five independent random-number streams with
means from $-206.17$ to $-194.10$ (overall $-198.82$). By contrast, confidence-ranked arbitration
returns $-443.37$; mean-reward-first, support-first, defer-to-ensemble, and unanimous-only
arbitration all return $-500$; and the single-source fallbacks return $-499.5$ and $-500$. A matched
control that ignores the rules entirely and samples uniformly random actions averages $-498.71$
across five replicates. Confidence-ranked arbitration is therefore not merely suboptimal on this task
but \emph{anti-selective under the evaluated conflict and occupancy regime}: choosing rules by
confidence selected worse actions than choosing them at random. This ordering is an artifact of the
induction protocol: the bank was induced from \emph{sampled} training actions but deployed under
\emph{argmax}. Under a protocol-consistent, argmax-consistent (greedy) bank the same confidence-ranked
arbiter reaches $-141.77$ and \emph{beats} random arbitration ($-234.26$); see
Section~\ref{sec:experiments:acrobot:protocol}.

A 200-state Monte-Carlo action-semantics probe (fresh states from deterministic actor rollouts,
Table~\ref{tab:mcprobe}) was historically used to dissect the failure. It is \emph{not} a valid
diagnostic of the deployed 8-actor fusion policy: under A2's matched 8-actor continuation, every
forced first action returns $-500$ and the Monte-Carlo estimator is completely degenerate. The
historical 2-actor probe reported only 22.5\% agreement between the confidence-ranked action and the
best first action (counterfactual continuation return under the actor-logit ensemble), and the fusion
policy conflicts on 90.5\% of the 200 states with an action histogram sharply different from the
Monte-Carlo-optimal one; but these figures characterize the 2-actor probe and do not transfer to the
deployed fusion. The mean margin between the best and second-best first actions under that probe is
108.24 return units, so the decisions are not near-ties.

\begin{table}[t]
\centering
\caption{Acrobot-v1: arbitration ablations on the 100 matched held-out episodes. Paired intervals
are bootstrap 95\% intervals of the per-episode difference relative to the current
confidence-ranked policy.}
\label{tab:ablations}
\begin{tabular}{lcc}
\toprule
Arbitration rule & Mean return & Paired 95\% interval vs.\ current \\
\midrule
Confidence-first (current) & $-443.37$ & --- \\
Confidence only (no tie-breaks) & $-444.49$ & $[-3.16,\ 0.06]$ \\
Random candidate & $-187.02$ & $[231.38,\ 278.21]$ \\
Mean reward first & $-500.0$ & $[-75.47,\ -39.67]$ \\
Support first & $-500.0$ & $[-74.84,\ -39.68]$ \\
Defer conflicts to ensemble & $-500.0$ & $[-75.12,\ -40.08]$ \\
Unanimous only & $-500.0$ & $[-74.49,\ -39.36]$ \\
Best single fallback & $-452.69$ & $[-19.51,\ 0.50]$ \\
Single source 11 / 29 & $-499.48$ / $-500.0$ & --- \\
\bottomrule
\end{tabular}
\end{table}

\begin{table}[t]
\centering
\caption{Acrobot-v1 Monte-Carlo action-semantics probe on 200 fresh states (state sample from new
deterministic actor rollouts). ``Best first action'' is the action maximizing counterfactual
continuation return when followed by the actor-logit ensemble. The probe forces a candidate first action
and follows a 2-actor mean-logit continuation; it characterizes the 2-actor probe and does not transfer
to the deployed 8-actor fusion (A2).}
\label{tab:mcprobe}
\begin{tabular}{lc}
\toprule
Quantity & Value \\
\midrule
Fusion coverage & 100\% \\
Fusion conflict rate & 90.5\% \\
Fusion agreement with any source-$Q$ argmax & 64.5\% \\
Fusion agreement with ensemble-GPI argmax & 0.0\% \\
\textbf{Fusion agreement with MC-best first action} & \textbf{22.5\%} \\
MC continuation return, first action 0 / 1 / 2 & $-132.9$ / $-202.5$ / $-157.1$ \\
MC-best first-action histogram (0/1/2) & 87 / 46 / 67 \\
Fusion action histogram (0/1/2) & 0 / 129 / 71 \\
Mean best-vs-second margin & 108.24 \\
\bottomrule
\end{tabular}
\end{table}

\subsection{Protocol-consistent induction recovers Acrobot fusion}
\label{sec:experiments:acrobot:protocol}

The Acrobot failure reported above is a train/deploy protocol artifact, substantially explained by the
induction/deployment mismatch in this recorded setup, while the remaining high conflict and any
independent task-complexity effect remain unresolved. The recorded numbers were all measured under
\emph{argmax} deployment, where
every base actor already sits at the $-500$ floor, using a rule bank induced from \emph{sampled}
training actions. A1b re-runs the composition under a $2\times2$ of induction protocol (sampled actions
vs.\ argmax-consistent greedy labels) and evaluation protocol (argmax vs.\ sampled), holding the same
recorded traces and reset seeds (the harness self-test reproduces $-443.37$ exactly).

\begin{table}[t]
\centering\small
\caption{Acrobot-v1 protocol-consistent composition (A1b). Mean held-out return over 100 episodes at
reset seeds $20\,000\,000+i$. Bank = rule-label source; Eval = action-selection rule at deployment.
The recorded config is the sampled bank under argmax.}
\label{tab:protocol}
\begin{tabular}{@{}l l c c@{}}
\toprule
Policy & Bank (induction) & argmax eval & sampled eval \\
\midrule
Best single actor (base) & --- & $-500.00$ & $-332.14$ \\
Actor-logit ensemble (base) & --- & $-500.00$ & $-229.34$ \\
Grammar fusion $F$ & sampled & $-443.37$ & $-436.87$ \\
Grammar fusion $F$ & greedy (protocol-consistent) & \textbf{$-141.77$} & $-133.21$ \\
Random candidate & sampled & $-187.02$ & $-186.69$ \\
Random candidate & greedy (protocol-consistent) & $-234.26$ & $-229.64$ \\
\bottomrule
\end{tabular}
\end{table}

Under the protocol-consistent greedy bank, confidence-ranked fusion reaches $-141.77$, \emph{beating}
random arbitration ($-234.26$) by $+92.49$ and the best argmax base actor ($-500.00$) by $+358.23$---the
best of all eight strategies tested. Coverage (95.4\% vs.\ 95.7\%) and conflict rate (87.6\% vs.\ 89.2\%)
are essentially unchanged between the two banks, so the $+301.60$ gain is the label-protocol
consistency itself, not additional coverage. The evaluation protocol barely matters ($\sim$8-point
swing) compared with the induction protocol ($\sim$301-point swing). On CartPole-v1 the sign flips:
the \emph{sampled} bank remains marginally better (205.96 vs.\ 191.41), so protocol consistency is
necessary on Acrobot but not a universal rule. Acrobot's recorded fusion failure is therefore a
methodology artifact (train/deploy mismatch), not evidence that more complex tasks cannot be fused;
whether task complexity imposes a \emph{separate} ceiling once induction is protocol-consistent is an
open question for future work.

\subsection{Behavioral agreement: rule overlap does not transfer}
\label{sec:experiments:agreement}

\begin{figure}[t]
\centering
\includegraphics[width=0.72\textwidth]{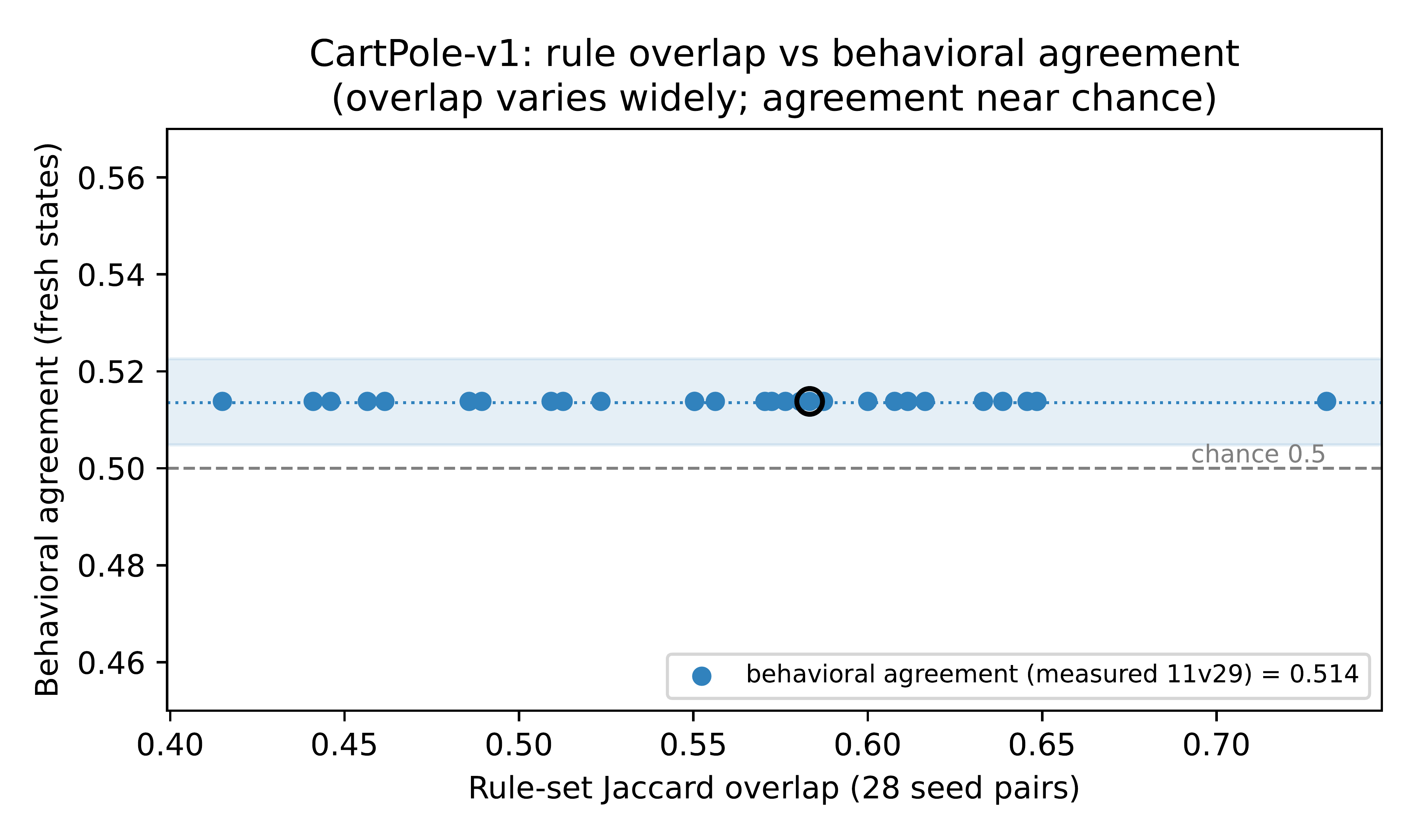}
\caption{CartPole-v1: rule-set overlap (Jaccard, seeds 11 vs.\ 29 and all 28 pairs) versus
behavioral agreement on 10{,}000 fresh states. Error bar: 95\% bootstrap interval of the
cluster-balanced agreement. The dashed line marks chance (0.5).}
\label{fig:agreement}
\end{figure}

Rule-set overlap and behavioral agreement measure different things, and the data show how different
they are (Figure~\ref{fig:agreement}). On CartPole-v1, the rule banks of seeds 11 and 29 share a
Jaccard overlap of 0.58 and agree on all 77 shared conditions, and the mean pairwise rule overlap
is 0.56. Yet on 10{,}000 fresh states (5{,}000 from each actor's own occupancy), the two policies
select the same action on only 51.38\% of states (episode-cluster-balanced estimate 51.36\%; 95\%
bootstrap interval $[50.48,52.26]$). Within each actor's own occupancy the agreement is 50.6\% and
52.2\% respectively. Stratifying by pole angle shows the agreement is near chance for near-zero
angles (50.2\%) and only modestly above chance in the mid range (55.8\%). The rules of these two
agents share a Jaccard overlap of only $0.58$, yet the agents behave almost independently on fresh states.
Rule-set overlap is therefore not a proxy for behavioral agreement; it is an upper-layer statistic
that must be validated against the behavior itself.

\clearpage

\subsection{Opportunity baseline: what a single rule bank already covers}
\label{sec:experiments:opportunity}

The coverage statistics reported above are occupancy-relative: $\mathrm{cov}$ is measured on the
steps a policy itself visits, so a higher value can mean either that the bank describes more of the
space or that the policy has been pushed into the region its own bank dominates. Both readings are
consistent with the same number, which is why coverage is usable as a descriptive statistic within
one arm but not as an outcome variable across arms (Section~\ref{sec:discussion:limitations}).

To separate the two readings we report the denominator that does not move with the policy: with all
eight rule banks \emph{frozen}, what fraction of the union of observed condition symbols does a
single bank accept? Table~\ref{tab:opportunity} reports the result. On CartPole-v1 the eight banks
admit $193$ of the $256$ grid cells ($75.4\%$), and a single bank already accepts
$59.7\%$ of those $193$ on average ($51.8\%$ to $75.1\%$). On Acrobot-v1 the eight banks admit
$608$ of $4096$ cells ($14.8\%$), and a single bank accepts $34.8\%$ on average
($12.7\%$ to $49.8\%$).

The consequence bounds what description-layer composition can buy. On CartPole-v1, drawing one seed's
bank at random already covers a majority of the observed vocabulary, so merging eight banks adds
descriptive reach slowly; on Acrobot-v1 the same quantity is a third. The sharper number is the common
core: the conditions accepted by \emph{all eight} banks are $38$ on CartPole-v1 and $2$ on
Acrobot-v1, and on Acrobot-v1 no condition in that core receives the same action from all eight
($0$ same-action core). The intersection---the only region where a description-layer arbiter has two
or more admissible candidates and can therefore change the emitted action---is a vanishing fraction of
the observed vocabulary in the conflict-dominated regime.

One comparison in the same table is a caution rather than a result. On shared conditions, two banks
emit the same action in $26$ of $28$ CartPole-v1 pairs, with mean agreement $0.999$. That near-perfect
value is a property of the four-bin quantile symbolizer, not evidence of a shared policy core: with
four bins per dimension the symbol is a coarse state ordering, and the induced action tracks it almost
deterministically. Agreement measured at the symbol level therefore has almost no discriminative
power, and any composition claim built on symbol-level overlap inherits this ceiling. Acrobot-v1 shows
the complementary case, where the same statistic falls to $0.450$ and one pair reaches perfect
agreement---the symbolizer is not the binding constraint there, conflict is.

\begin{table}[t]
\centering\small
\caption{Opportunity baseline over frozen rule banks. ``Observed vocab'' is the union of condition
symbols accepted by any of the eight seed banks; the theoretical grid is $4^{d}$ and is not used as a
denominator. ``Single-bank acceptance'' is the fraction of the observed vocabulary one bank accepts,
reported as mean over seeds with the range in parentheses. Computed by
\texttt{auditability/analyze\_opportunity\_baseline.py} from the frozen
\texttt{induced\_grammar.json} artifacts.}
\label{tab:opportunity}
\begin{tabular}{@{}l c c@{}}
\toprule
Quantity & CartPole-v1 & Acrobot-v1 \\
\midrule
State dimensions $d$ & 4 & 6 \\
Theoretical grid $4^{d}$ & 256 & 4096 \\
Observed vocab $|\mathcal{V}|$ & 193 & 608 \\
Grid cells admitted ($|\mathcal{V}|/4^{d}$) & 75.4\% & 14.8\% \\
Single-bank acceptance of $|\mathcal{V}|$ & 0.597 {\scriptsize(0.518--0.751)} & 0.348 {\scriptsize(0.127--0.498)} \\
Conditions accepted by all eight banks & 38 & 2 \\
{\quad}of which same-action across all eight & 38 & 0 \\
Bank pairs agreeing on shared conditions & 26/28 & 1/28 \\
\bottomrule
\end{tabular}
\end{table}

\subsection{Causal symbol intervention as a separate sub-study}
\label{sec:experiments:causal}

The communication experiment is reported apart from rule fusion because it answers a different question:
whether a \emph{learned discrete symbol} in a designed task causally influences a receiver's decision,
not whether state--action rule fusion composes policies. On the redesigned private-target game, six
independent runs of $100{,}000$ decisions each ($600{,}000$ total) give receiver accuracy $0.998923$
with the true message, $0.503292$ with a shuffled message, and $0.500$ with no message. The
matched-support intervention---replacing the message while holding the state fixed---changes the
receiver's action probability by $\bar{\Delta}=0.998366$ on average (range $0.996986$--$0.999685$
across runs), with common-support overlap $0.94308$. In the seed-0 probe, per-position total-variation distance between
even/odd token marginals is $0.0967$ at position~0 and $0.9984$ at position~1, confirming the
information-bearing position identified by the per-position MI analysis.

Under the controlled intervention protocol, which holds the receiver state and task context fixed to
the extent implemented by the environment, we estimate the effect of replacing the discrete message on
the receiver's action probability; the claim is internal to this designed task and depends on the
stated common-support and no-private-channel checks. Its scope is deliberately
narrow: it shows that \emph{in this designed task} the learned discrete symbols causally influence the
receiver's decisions. It does \emph{not} show that the symbols are human-interpretable, compositional,
or ``meaningful'' in any richer sense, and it does not generalize to other RL agents. All causal claims
in this paper rest on the redesigned task and the staged gate of Section~\ref{sec:method:causal}; the
original parity game is reported only as a negative result.

\subsection{Fitted-Q GPI baseline failure}
\label{sec:experiments:fqe}

The fitted-Q GPI baseline (\ref{eq:fqe}--\ref{eq:gpi}) fails in both environments, and the failure
mode is informative. On CartPole-v1 its mean return is 9.34 (paired difference $-154.31$ versus the
best single actor), and the fitted critics are nearly flat: the critic values have cross-seed
agreement of 0.84 on greedy actions, but their pooled $Q$-value distribution has a standard deviation
of only 0.81 around a mean of 10.7, i.e., the critics vary little across states---a
calibration/composition failure rather than a usable value signal. The FQE-GPI action
histogram is 30{,}758:805 (nearly always action 0), and its agreement with the actor-logit ensemble
on 31{,}563 held-out states is 35.3\%. On Acrobot-v1 the collapse is total: FQE-GPI selects action 0
on all 50{,}000 held-out states, the critics are constant at $-11.44$ with standard deviation
0.009, greedy-action agreement between critics is 6.3\%, and the held-out Bellman residual is 0.92
(MAE), which is far from the precision that the GPI bound would require. These are diagnostics, not
discoveries: they show that fitted-Q composition over a 200k-transition reservoir is unreliable in
these setups, and they caution against claiming that the rule-fusion result is ``better than
value-based composition'' on the strength of this baseline alone.

\subsection{Lossless coding results}
\label{sec:experiments:codec}

The codec of Section~\ref{sec:method:codec} is exact by construction (the decoded payload hashes to the
input hash). On CartPole-v1 it processes $9{,}178{,}724$ tokens across $16$ grammar productions; the raw
trace is $61.3$\,MB, gzip compresses it to $12.3$\,MB, and the grammar--gzip archive is $13.4$\,MB. On
Acrobot-v1 the corresponding figures are $8{,}549{,}496$ tokens, $57.6$\,MB raw, $12.5$\,MB gzip, and
$13.7$\,MB grammar--gzip. The two-part code length therefore \emph{does not} beat generic compression:
the grammar adds structure without removing bytes, and the grammar--gzip archive is $8.4\%$ larger than
gzip on CartPole-v1. The two-part code length $L_2(\tau)$ (Eq.~\ref{eq:twopart_code}), expressed as a
ratio to the uncompressed raw trace, sits at $0.51$--$0.58$ across the six 50-episode prefixes of a
separate seed-11 CartPole probe; these prefix-level two-part code lengths come from that single seed-11
probe and
not from the full eight-seed archive. The two-part code is shorter than the uncompressed reference, but
the stored grammar archive is not smaller than gzip. The result is reported honestly as a fidelity tool,
not a storage win: the codec earns
its place by enabling exact replay-against-compressed-form checks, not by shrinking the log.

\subsection{Summary of verdicts}
\label{sec:experiments:verdicts}

\begin{table}[t!]
\centering\small
\setlength{\tabcolsep}{4pt}
\caption{Summary of measurement layers. Identification-bearing results are distinguished from
diagnostics and operational-health checks. The two arbitration rows state which comparator is better
in each bank regime; each cell is a paired comparison, not a return.}
\label{tab:verdicts}
\begin{tabular*}{\textwidth}{@{\extracolsep{\fill}}lccc@{}}
\toprule
Criterion & Layer & CartPole-v1 & Acrobot-v1 \\
\midrule
Trace/hash/replay validity & operational health & pass & pass \\
Rule overlap across seeds & diagnostic & 0.557 & 0.096 \\
Fusion vs.\ best-training actor (paired $\Delta$) & measured, capped & $+42.31$ & $+56.63$ \\
Fusion vs.\ best held-out actor (paired $\Delta$) & measured, capped & $-294.04$ & $+56.63^{\dagger}$ \\
Fusion mean return & measured & 205.96 & $-443.37$ \\
Arbitration winner (sampled bank) & identification & n/a & random, $-187.02$ \\
Arbitration winner (greedy bank) & identification & n/a & confidence, $-141.77$ \\
Behavioral agreement (fresh states) & identification & 51.4\% ($\approx$ chance) & n/a \\
FQE-GPI mean return & diagnostic & 9.34 & $-499.57$ \\
\bottomrule
\end{tabular*}
\end{table}

$^{\dagger}$On Acrobot-v1 all eight source actors return $-500.0$ on the matched set, so the best
held-out actor and the post-hoc best-training actor coincide and both give $+56.63$; on CartPole-v1
they differ sharply, which is the point of the row.

Table~\ref{tab:verdicts} separates the layers. The identification-bearing findings are (i) the
paired fusion advantage on CartPole-v1 against the post-hoc-selected comparator---reported together
with its reversal against the best held-out actor, so that the row is not read as a general
performance claim, (ii) the behavioral-agreement near-chance result, and
(iii) the Acrobot arbitration reversal. All other rows are diagnostics or health checks.

% ----------------------------------------------------------------------
\section{Discussion}
\label{sec:discussion}

\subsection{Why the mechanism may work, and where it breaks}
\label{sec:discussion:mechanism}

The CartPole advantage of rule-level fusion has a simple account in the equations. The fused
decision (\ref{eq:fusion}) selects the highest-confidence rule among the candidates
(\ref{eq:admission}); when conflicts are rare (0 of 77 shared conditions), the fused policy is
essentially a union of the seeds' agreement, and the fallback (\ref{eq:fusion}, second case) covers
the residual 1.7\% of states. The paired comparison controls for the reset-seed distribution, and
the bootstrap interval excludes zero. But the same equations fail on Acrobot-v1 because the
confidence statistic $f_{\mathbf{c},a}$ (\ref{eq:rulestats}) is a \emph{within-trace} conditional
frequency, not a predictor of which action is best in a new state: when 89\% of decisions conflict
and the best first action (probe, Table~\ref{tab:mcprobe}) disagrees with the fused choice on
77.5\% of states, the arbitration is selecting on a signal that does not rank actions by return.
The reversal---random arbitration beating confidence arbitration---is consistent with the
Monte-Carlo probe: in this Acrobot probe the confidence-ranked action agreed with the estimated best
first action on only 22.5\% of the 200 probe states, which is below the 26.7\% expected if the fusion
action histogram and the Monte-Carlo-best histogram were independent, so removing the signal helps.
We therefore treat this as an induction/deployment protocol artifact rather than evidence that
confidence is generally anti-selective. A1b identifies the cause directly: the bank was induced from
\emph{sampled} training actions but deployed under \emph{argmax}, where all base actors sit at the
$-500$ floor; re-inducing from argmax-consistent (greedy) labels under the same deployment raises
confidence-ranked fusion to $-141.77$ and \emph{reverses} the ordering (random arbitration then returns
$-234.26$). The 22.5\% Monte-Carlo agreement figure (Table~\ref{tab:mcprobe}) is a 2-actor continuation
probe (A2) that does not transfer to the 8-actor fusion used in deployment, so it should not be read as
a property of the deployed arbiter.

\subsection{Prediction outcomes}
\label{sec:discussion:predictions}

Two expectations held and two failed. Expected: (i) trace fidelity (replay, hashes) would verify
cleanly, and (ii) rule overlap would be higher under a shared symbolizer than under per-seed
symbolizers (the design motivation). Unexpected: (iii) behavioral agreement would track rule
overlap---it does not (51.4\% vs.\ chance), and (iv) confidence-ranked arbitration would dominate
random arbitration---it is reversed on Acrobot-v1. For (iii), an alternative explanation is that
the shared symbolizer is too coarse: with only four bins per dimension, two policies can agree on
rules for the symbol while disagreeing on nearly every state inside the symbol; the rules describe
the \emph{symbol}, and behavior lives on \emph{states}. For (iv), the reversal is now resolved by A1b as an induction/deployment protocol artifact: under
protocol-consistent greedy-bank induction, confidence-ranked fusion ($-141.77$) beats random arbitration
($-234.26$), so the confidence signal is not generally anti-selective---the recorded reversal came from
the sampled-bank labels deployed under argmax. The coarse-symbolizer alternative for (iii) remains open;
the protocol-mismatch explanation for (iv) is established.

\paragraph{Occupancy action histograms as one mechanism for near-chance agreement.}
The behavioral-agreement result has a concrete mechanistic reading. On the 5{,}000 states drawn from
seed~29's held-out occupancy, seed~11 selects action~0 on $4994/5000$ states, whereas seed~29 itself
splits nearly evenly ($2604$ action~0 versus $2396$ action~1); the asymmetry inverts on seed~11's
occupancy, where seed~29 selects action~1 on all $5000$ states and seed~11 splits $2472/2528$. Each
actor is therefore close to deterministic inside the other's occupancy, and the two opposite,
near-uniform action preferences yield a pooled agreement near the binary chance baseline. We report
this as a mechanism \emph{consistent with} occupancy asymmetry as one explanation; we do not claim it
is the unique cause, and the coarse four-bin symbolizer remains a candidate confounder.

\subsection{What the protocol does \emph{not} establish}
\label{sec:discussion:not}

The positive results are bounded, and the boundaries are the point. Specifically:
\emph{(i)}~rule overlap does not imply behavioural agreement (Section~\ref{sec:experiments:agreement});
\emph{(ii)}~trace replay does not imply causal explanation, and re-executing recorded transitions
recovers the record but not the optimizer state or random-number stream that produced it;
\emph{(iii)}~provenance visibility does not imply recovery of the neural mechanism;
\emph{(iv)}~the fitted-Q baseline failing its own gate does \emph{not} imply that rule fusion is better
than value-based composition---only that the comparator is unqualified here;
\emph{(v)}~the random-arbitration anomaly does \emph{not} imply that a random arbiter is the generally
optimal resolver. A plain actor-logit ensemble also outperforms rule fusion in CartPole ($500.00$ vs
$205.96$) while being simpler, and so does the population it summarises: on the matched episodes the
mean and median individual actor reach $330.86$ and $358.81$, and four of the eight actors reach the
ceiling (Table~\ref{tab:comparators}). The objection is correct about performance and misses the point
about
auditability. The ensemble cannot say \emph{which} knowledge produced a decision, cannot report that two
sources disagreed, and cannot declare a blind spot. What fusion buys over the ensemble is not return but
\emph{inspectability per decision}; what the hash chain buys over a log file is not accuracy but
\emph{checkability}. Whether that tradeoff is worth it depends on the deployment; the claim is only that
the tradeoff exists and can be measured honestly.

\subsection{Same framework, task-specific regimes}
\label{sec:discussion:regimes}

CartPole-v1 and Acrobot-v1 are fused with the \emph{same} provenance-preserving rule-level
fusion framework. Both environments follow the identical induction-and-arbitration principle:
(1)~each actor produces state$\to$action rules; (2)~multiple actors contribute candidate rules
for the same state; (3)~when the candidates agree on the action, the common action is taken;
(4)~when they conflict, the action is chosen by a ranker over confidence, support, and reward;
(5)~when no rule covers the state, a blind-spot fallback actor is used; and (6)~the rule source
and the full decision provenance are recorded for every step. The two tasks therefore do
\emph{not} use two different fusion methods.

What differs is everything task-specific around that shared skeleton
(Table~\ref{tab:tasks}). State dimension (4 vs.\ 6), action count (2 vs.\ 3), actor behavior
distributions, rule-bank sizes (about 921/1218 sampled/greedy rules on CartPole vs.\
1694/2434 on Acrobot), coverage, and conflict rates all differ, so the realized fusion behavior
and the diagnostics that matter are not directly comparable.

\begin{table}[t]
\centering\small
\caption{CartPole-v1 vs.\ Acrobot-v1 under the same rule-fusion framework. Fusion returns are
mean held-out return over 100 episodes (sampled bank = original induction, greedy bank =
protocol-consistent induction).}
\label{tab:tasks}
\begin{tabular}{@{}l c c@{}}
\toprule
Aspect & CartPole-v1 & Acrobot-v1 \\
\midrule
State dimensions & 4 & 6 \\
Actions & 2 & 3 \\
Rule-bank size (sampled / greedy) & 921 / 1218 & 1694 / 2434 \\
Sampled-bank fusion return & 205.96 & $-443.37$ \\
Greedy-bank fusion return & 191.41 & $-141.77$ \\
Conflict level & very low (0.4\%) & very high (89.2\%) \\
Protocol-mismatch impact & relatively small & very large \\
MC ranker status & not fully validated & 8-actor version degenerates to all action 0 \\
\bottomrule
\end{tabular}
\end{table}

The same provenance-preserving rule-fusion framework was applied to both tasks, while
task-specific action spaces, actor populations, rule-bank construction protocols, and
arbitration diagnostics differed.

Crucially, Acrobot's improved number is \emph{not} a different fusion algorithm. It is the same
confidence-ranked fusion with blind-spot fallback, applied to a rule bank rebuilt under the
same greedy (argmax-consistent) protocol used at deployment. The change corrects an
\emph{experiment protocol}---induction and deployment now agree---rather than altering the
fusion principle. The decisive difference between the two tasks is therefore not ``two fusion
methods'' but two regimes. On CartPole the sampled and greedy banks differ little and the open
question is whether fusion beats the best single actor; on Acrobot the sampled/argmax mismatch
severely distorts the result, and rebuilding the bank under a consistent protocol recovers
fusion from $-443.37$ to $-141.77$ (Section~\ref{sec:experiments:acrobot:protocol}).

\subsection{Theoretical and practical implications}
\label{sec:discussion:implications}

Theoretical implication: under a shared, frozen symbolizer, rule-level composition inherits
provable properties only for the statistics that define it (\ref{eq:rulestats}--\ref{eq:admission});
no guarantee about held-out behavior follows from rule overlap, and none should be claimed. The
behavioral-agreement result is direct evidence for this boundary. Practical implication: rule
fusion is usable as an \emph{audit and composition} layer in low-conflict regimes, and is not a
reliable training-time constraint (Section~\ref{sec:evolution:constraint}): that intervention is
sign-inconsistent across seeds and damages the seeds that already reach the ceiling. We therefore
make no claim that fusion trains better policies, and we note that the constrained arm's effect is
heterogeneous rather than uniformly negative.

\paragraph{Auditability as a vector, not a score.}
The six predicates of Section~\ref{sec:intro:props} should be reported as a vector, never averaged into
one number. The reason is empirical: in this study the components point in different directions---trace
integrity passes cleanly while behavioural agreement fails, and composition improves on CartPole but
collapses into conflict on Acrobot. A single aggregate would conceal exactly the heterogeneity that the
failure-driven history was built to expose. We therefore define
\[
\begin{aligned}
\mathrm{Auditability} = \big(&\text{trace integrity},\ \text{lossless coding},\ \text{rule coverage},\\
&\text{behavioural agreement},\ \text{composition quality},\ \text{value-model reliability}\big),
\end{aligned}
\]
and treat any claim that averages them as out of scope. This is the same six-item list as
Table~\ref{tab:props}, and the predicates are fixed before the results are known; the
verification mechanisms of Section~\ref{sec:method:ledger} (hash-chain integrity, environment replay,
provenance resolution) supply evidence for selected predicates without being predicates themselves.
In particular, replay fidelity and provenance visibility are not interchangeable with composition
quality and must not be substituted for it in the vector.

\subsection{Limitations}
\label{sec:discussion:limitations}

The comparison is bounded in several ways. (1) The seed pool is $n=8$; we report cross-episode
uncertainty, not cross-seed uncertainty. (2) The single-actor baseline and the single-actor
fallback ablation select an actor by training return mean, which is a post-hoc choice; on
CartPole-v1 the selected seed (43) scores 163.65 on the matched evaluation set while four of the other
seven seeds reach the
500 ceiling there, so the selection basis is demonstrably weak,
and a different selection rule reverses the sign of the comparison
(Table~\ref{tab:comparators}). (3) Environment ceilings make
the CartPole comparison partially saturated (four of eight baselines already reach 500). (4) The
confidence threshold and the arbitration rule were chosen after observing earlier failures; the
two-setting threshold exploration is exploratory, not a systematic search. (5) The FQE-GPI baseline
is approximate and off-policy; its failure is a diagnostic, not a general statement about GPI.
(6) The behavior-agreement sample draws states from each actor's own occupancy, so the pooled
agreement is occupancy-weighted, not uniform over the state space. (7) Acrobot-v1's random
arbitration advantage was replicated over five random streams but the mechanism remains open.
We disclose each of these rather than repairing them by assumption.

\paragraph{Representation and task-design limitations.}
The shared quantile symbolizer is itself a confounder: with only four bins per dimension, two policies
can agree on a rule for the symbol while disagreeing on nearly every state inside it, so the rules
describe the \emph{symbol} and behavior lives on \emph{states}. The communication intervention is a
designed task and does not transfer to general RL agents; an early payoff design made communication
unnecessary, a continuous encoder path could bypass the discrete codebook, and the blind receiver
control had to truly block private information. The value baseline's quality gate had to be pre-registered
precisely because it can fail silently.

\paragraph{Future work: from description-layer arbitration to generative composition.}
The composition studied in this paper operates on the \emph{description} layer: the fused policy
jointly executes rules admitted from a common task, and its reachable behavior is bounded by the
coverage of the component rule banks. The opportunity baseline of
Section~\ref{sec:experiments:opportunity} makes that bound concrete---a single bank already accepts
$0.597$ of the observed symbol vocabulary on CartPole-v1 ($0.348$ on Acrobot-v1), and the eight-bank
common core is $38$ and $2$ conditions respectively---so merging banks adds descriptive reach slowly.
A different scientific object is \emph{generative} composition: several independently trained
skills---locomotion, balance, perturbation recovery, spin control---each a policy for its own sub-MDP,
are composed by the system itself into a qualitatively new behavior (e.g., a backflip) whose
trajectories visit states outside every component's coverage. Each stage below is the next question
raised by a specific result in this paper, rather than an independent research agenda, and we state for
each one what it inherits from the reported findings; none of the stages is solved here, and each is
stated with a gate that can fail.

Stage~1 (\emph{temporal abstraction}). The present description layer exposes no timing structure: a
rule commits to one action for the current state and nothing more, so multi-phase behavior cannot be
represented. The gap is demonstrated rather than asserted, because the audited artifacts carry no
timing field to recover: two banks can be replayed and audited at the rule level while neither one
contains a termination time, so ``rule replay succeeds'' and ``a complete skill is represented'' are
different predicates---the same non-implication the rest of this paper measures. Concretely, on Acrobot-v1
a single rule registers up to $12{,}953$ supporting transitions, which shows that a rule can be
heavily exercised without carrying any notion of when its action should stop. Wrap each skill's rule
bank in an option $\omega=(I_\omega,\pi_\omega,\beta_\omega)$
consisting of an initiation set, an intra-option policy built from the induced rules, and a
termination condition. The component policy remains the executable low-level controller for the
switching baseline; the substantive requirement is not that the option run for $k$ steps but that
$\beta_\omega$ remain reliable under the distribution \emph{induced by the option itself}, which is
the same self-confirmation hazard that governs rule re-induction here---a skill that looks stable only
inside the narrow occupancy it creates for itself is not a composable skill. Gate: per-option replay
and behavioural agreement; the distribution of option duration; the blind-spot rate inside
$I_\omega$; the safety-envelope violation rate; and acceptance reported two ways, as a
probe-weighted fraction over an independently fixed probe set and as a fraction of the observed symbol
vocabulary, never as one number alone.

Stage~2 (\emph{cross-task skill interfaces}). This stage follows directly from the
behavioral-agreement null result, generalized from two policies to two skills. If two rule banks
overlapping on $77$ shared conditions with zero conflicting conditions still disagree on $48.6\%$ of
fresh states (Section~\ref{sec:experiments:agreement}), then overlap between two \emph{skills} cannot
be read as evidence that one can hand control to the other. Options define \emph{when} to switch, not
\emph{how to superpose}. Pure switching is therefore a legitimate baseline, but it cannot by construction exceed
the safe envelope of the component control set, which is why the baseline is a baseline and not one
candidate among equals. Generation requires an additional composition operator outside the options
framework---a residual, mixture, or re-parameterised action channel---whose parameters are selected by
the level above, so that the high level chooses \emph{when} to engage the operator rather than
emitting motor actions directly. Alongside this, extend the frozen calibration symbolizer to a shared
abstraction across sub-MDPs, which requires shared state and action semantics that the present
per-task symbolizer does not assume, and declare entry/exit compatibility: option $B$'s initiation set
should overlap option $A$'s termination set so that skills can chain. Compatibility must be verified in
\emph{state} space rather than symbol space. The reason is in Table~\ref{tab:opportunity}: two banks
agree on $26$ of $28$ CartPole-v1 pairs at the symbol level with mean agreement $0.999$, a value set by
the coarse quantizer rather than by the policies, so symbol-level overlap has almost no discriminative
power. Gate: chain reachability---a composed rollout whose states lie outside the union of component
coverage---with transition success, safety margin, an uncertainty bound, and validation on a probe set
that the composed rollout did not generate.

Stage~3 (\emph{composition search}, the generation layer). The present fusion is an arbiter, not a
generator: it selects among suggestions that already exist, $F(s)=\arg\max$ over the admissible
candidates of one state (Eq.~\ref{eq:fusion}). It therefore inherits two properties this paper
measured and could not repair. First, its correctness is bounded by the admission protocol that built
the bank, not by the knowledge in it: Acrobot-v1 moves from $-443.37$ to $-141.77$ purely by
re-inducing the bank from argmax-consistent labels
(Section~\ref{sec:experiments:acrobot:protocol}), so an arbiter can fail without any component being
wrong. Second, on conflict steps the arbiter appeared no better than an honest baseline: it disagreed
with the eight-actor logit ensemble on $50.6\%$ of conflict steps (Section~\ref{sec:discussion:predictions}).
These two facts are what make \emph{search} the successor object rather than a better ranking rule.
Train a high-level policy over the
skill-and-operator space against a sparse goal predicate, so that any novelty is attributable to the
optimizer's search over option sequences rather than to human-designed choreography. This is the
declared bottleneck, because the components must be selected under an induction protocol that has to
be fixed \emph{before} search begins rather than tuned in response to its outcome. Gate: the novelty,
attribution, and achievement checks of Stage~4, together with
a comparison against the switching baseline, single-skill, fixed-sequence, random-sequence, greedy, and
learned-planner arms.

Stage~4 (\emph{emergence audit}, a seventh predicate). This stage extends the paper's central
discipline---one predicate passing does not license another (Section~\ref{sec:intro:props})---from six
predicates over a single policy to a composed one, and it is where that discipline becomes
load-bearing rather than expository: a single successful rollout is exactly the kind of evidence this
paper refuses elsewhere, since trace integrity can pass while behavioural agreement fails, and a bank
can raise its own coverage while its agreement with the network falls
(Section~\ref{sec:discussion:limitations}). Make ``novel behaviour'' an auditable property
rather than a claim, with the criteria pre-registered before Stage~3 runs and measured after it. Five
checks are reported separately and none is promoted to a single novelty score, because this paper's own
results are the counterexample to any single number: the six predicates of
Section~\ref{sec:intro:props} disagree with each other here in practice, and an aggregate would have
concealed precisely that.
\emph{(i) State-level}: the fraction of composed-rollout states outside every component's acceptance
region, $\hat p_{\mathrm{outside}}$, must meet a declared threshold, with the construction rule of that
region, the outside count, the probe denominator, and a confidence interval all reported; a bare
``greater than zero'' is not a test, since one outlier from dynamics variation would satisfy it. The
region must be fixed before the composed rollout is inspected, and its denominator must be chosen
rather than inherited from the policy's own occupancy.
\emph{(ii) Control attribution}: on matched conflict steps, the composed action must deviate from the
honest eight-actor logit-ensemble prior. This is a \emph{non-reproduction diagnostic}---it rules out
the composition being a copy of the average actor prior---and it is explicitly not proof of a new
skill; it is reported with its matched denominator, an effect size, and a statistical test, and its
threshold is anchored to the measured conflict-step disagreement rate rather than to $1.0$, since that
baseline is already close to chance and agreement with it is therefore uninformative.
\emph{(iii) Achievement}: the composed planner must exceed fixed-sequence and random-sequence
baselines with the single-skill baseline unable to complete the goal, where the fixed-sequence
baseline is selected without reference to the composed result and is \emph{not} required to fail
universally---a hand-found sequence that happens to succeed is a valid solution, not a refutation.
This last asymmetry is the same one this paper applies to its own positive result, where the comparator
was selected post hoc by training return and the comparison was therefore reported as exploratory
rather than as a declared win.
\emph{(iv) Reproducibility}: the result must hold on reset seeds disjoint from those used to tune the
composition search, since the two-setting threshold exploration in this study was labelled exploratory
for exactly this reason (Section~\ref{sec:discussion:limitations}).
\emph{(v) Provenance}: hash-bound replay under the ledger protocol
(Definition~\ref{def:audit}, Eq.~\ref{eq:replay}) extended from single steps to option boundaries,
which is where the present results stop, together with a resolved source for every transition and an
explicitly declared construction for those that are novel. Note that a claim of novelty is a claim to
be \emph{audited}, not a licence from audit: five checks passing simultaneously is what would make such
a claim both new and checkable, and the schema extension in (v) is a prerequisite rather than a
formality.

\paragraph{Why the audit layer is enabling rather than sufficient.}
This extension does not claim that auditability alone produces a new behaviour; reinforcement
learning can discover isolated complex behaviour with no audit layer at all. The claim is narrower and
is about \emph{attribution} rather than discovery. In a single-task, single-policy setting an
aggregate return is sufficient evidence that something was learned. Under composition it is not: the
same return can be produced by a genuinely new skill sequence, by a component policy that was already
competent on the target behaviour, by an unsafe superposition of two skills, or by an
induction/deployment mismatch of the kind documented in
Section~\ref{sec:experiments:acrobot:protocol}, where the recorded fusion result inverts entirely
under a protocol correction ($-443.37$ versus $-141.77$). Return statistics cannot separate these
cases, because they do not record which knowledge produced which decision---the same reason the
actor-logit ensemble, which outperforms rule fusion on CartPole-v1 ($500.00$ versus $205.96$), cannot
report that two sources disagreed or declare a blind spot. What the rule-level layer adds is
attribution: per-decision provenance plus an independently defined acceptance region for each
component, which is what makes it possible to ask whether a composed behaviour is genuinely novel
rather than an accidental success, a hidden overlap, or a protocol artifact. Auditability is thus not
the skill-learning algorithm; it is the verification substrate that such an algorithm would need for
its output to be checkable. The present work establishes that substrate for a shared-task setting
only, and extending it to temporally extended, cross-task composition remains open.

\paragraph{Bounded expectation.}
The hard bottleneck is Stage~3: non-convergence of the composition search is itself a diagnostic---
evidence about the difficulty of the \emph{composition problem}, not about the intrinsic difficulty of
the target behaviour---provided that the induction and deployment protocols are first made consistent
for each component, since on Acrobot-v1 the recorded failure inverts entirely under a protocol
mismatch ($-443.37$ under the original sampled bank versus $-141.77$ under protocol-consistent
induction). We make no claim that any stage succeeds on current methods.

The unresolved mechanisms dictate the nearer-term next steps: the random-arbitration advantage is now
largely explained as an induction/deployment protocol artifact (A1b), so the remaining question is
whether \emph{task complexity} imposes a further ceiling once induction is protocol-consistent; test
return-weighted rule-bank induction under a pre-specified normalization and held-out evaluation; use
the native sequence symbols as the control representation; and evaluate behavior on non-saturated tasks
and across hardware.
The representation choice also separates this work from the ``AI Mother Tongue'' line: our
symbolizer is exogenous and frozen, whereas that line studies endogenous learned symbols
\cite{liu2025aim}, and the symbol-level agreement ceiling in Table~\ref{tab:opportunity} is a direct
consequence of that freezing. Whether a learned codebook can be made replay-verifiable by freezing it
per run and binding its assignments into the hash-bound ledger remains untested. This is a future
bridge, not a property established by the present experiments.

\paragraph{Material gaps (honest disclosure).}
The replay and hash checks are operational-health assertions on the recorded runs.
Cross-hardware replay is untested, the mechanism behind random arbitration remains undecomposed,
and the communication result is bounded to the designed task.

\subsection{Steelman}
\label{sec:discussion:steelman}

A $+42.31$ gain on a task where half the baselines already hit the
ceiling---against a comparator chosen to be the weakest-looking one---produced by a fusion policy that
is itself far below the ceiling and below the population mean, resting on one
fallback-seed choice and an arbitration rule that is catastrophically worse than random on the
second environment \emph{under the sampled-bank protocol}---this does not establish that rule-level
fusion is a good idea. We concede the
weight of this objection and answer it with the evidence rather than by inflating the claim. The
paper's primary contribution is not the magnitude of the CartPole gain; it is (a) a protocol in
which every rule is anchored to a replayed, hash-bound trace, so the description layer is
operationally verified against the recorded artifact; (b) direct evidence that rule overlap does not imply behavioral agreement, which
invalidates a common shortcut in the interpretability literature; and (c) an explicit, replicated
negative result---confidence-ranked arbitration can be anti-selective under the evaluated conflict and
occupancy regime. The CartPole gain is real but bounded (paired, bootstrap
interval excluding zero, matched episodes, post-hoc-selected comparator), and the Acrobot failure is
reported in the same table
that reports the gain, with the same measurement discipline. If the standard is ``auditable,
bounded, falsifiable evidence about when rule composition helps,'' the paper meets it.

\paragraph{Steelman, continued.}
The response is evidence, not framing.
First, the protocol record \emph{is} the claim: the paper promises exactly replayable,
provenance-visible composition under stated conditions and delivers the artifacts and the refutations
that bound it. Second, the random-rule finding \emph{constrains} rather than voids rule content: it is
separated from action noise by a matched control (random-candidate $-187.02$ versus uniform-random
action $-498.71$), which provides evidence that the candidate sets retain information beyond an
unconditional uniform-random-action baseline---although that control does not isolate whether the
information arises from rule content, state occupancy, action-frequency structure, candidate-set size,
or the selection procedure---even though the tested
deterministic resolver fails to exploit it. Third, the coordination task is disclosed as designed and
decision-relevant by construction; its evidential role is to show that a causal-symbol test can pass
\emph{somewhere} with proper gates, not that symbols are generally causal. 

\subsection{Replicability note}
\label{sec:discussion:replication}

All numbers above come from executed runs with hash-verified traces. The deterministic checks
(replay, hashes, codec round-trip) are expected to reproduce exactly; the diagnostic statistics
up to numerical precision; and the stochastic results (bootstrap intervals, random arbitration
replicates) within the reported intervals. Replaying a recorded output is not the same
as generating a new one; the random-arbitration replicates are fresh generations under fixed
streams.

\paragraph{Likely replication failures and their ratings.}
Three are foreseeable. \emph{(i)}~The arbitration result is sensitive to the selector stream: the initial
stream gave $-187.02$ while five later streams ranged to $-206.17$, so a single-stream replication would
report a different magnitude. \emph{(ii)}~The Acrobot result depends on the return floor: with a median of
$-500.00$, a replication with different reset seeds could shift the mean without changing any qualitative
conclusion. \emph{(iii)}~The behavioural-agreement result depends on the symbolizer, which is fitted to
random-policy occupancy; a replication with a different calibration policy could move it.

\emph{Evidenced mitigations.} Sharing one frozen symbolizer across seeds and arms within a task is
\textsc{high}---it removes the largest known confound in cross-seed comparison and is directly evidenced
by the boundary differences of up to $0.157$ observation units under per-seed fitting. Pairing all policy
comparisons on identical reset seeds is \textsc{high}---it is what makes the paired differences
interpretable given the return variance. Reporting the decision mix alongside every fused return is
\textsc{medium}---it prevents the most likely misreading. Repeating the random-selection condition across
five independent streams is \textsc{medium}---it establishes the effect is not a single-stream artifact.
For independent replication, the minimum requirement is the same frozen symbolizer, the same
admitted rule banks with their support and confidence statistics, and the same evaluation
reset-seed ranges.

% ----------------------------------------------------------------------
\section*{Reproducibility Statement}
\label{sec:repro}

All experiments use fixed seeds (control: 11, 29, 43, 71, 101, 149, 211, 307; composition comparisons:
100 matched reset seeds $20{,}000{,}000+i$; communication: 6 independent runs). Rule induction uses 4
quantile bins per state dimension under a shared symbolizer, minimum support 8, minimum confidence
$0.70$. Composition comparisons are matched on reset seeds. Because deep RL results can vary
substantially with random seeds, environment nondeterminism, evaluation protocol, and reporting
choices \cite{henderson2018deep}, we report seed identities, matched reset seeds, evaluation
distributions, and uncertainty intervals separately. The hash-chained trace logs, replay
validator, and tamper-probe results are archived with the experiment records. Analyses not pre-registered
are labeled exploratory. No human subjects, annotators, or sensitive data were involved.

\paragraph{Code and data availability.}
The experiment code, the compact hashed result package, and the manuscript sources are released in a
public repository at \url{https://github.com/cyrilliu1974/AuditableRL}. The full training traces,
optimizer-update ledgers, actor and training checkpoints, shared calibration metadata, induced rule
banks, and fitted-critic checkpoints (584 files, 4.7\,GB uncompressed) are deposited as an archival
dataset at \cite{liu2026auditablerl}. The dataset archive is accompanied by a SHA-256 checksum file
(\texttt{b4512466\ldots b4786}) so that a downloaded copy can be verified byte-for-byte before use;
unpacking the archive at the repository root restores the \texttt{runs/} layout that the released
reproduction commands expect. The compact result package records, for each headline number, the
source artifact path, byte count, and SHA-256, so every reported value can be traced to the
archived file that produced it without re-running training.

\section{Conclusion}
\label{sec:conclusion}

We converted the question ``what did the trained agents learn?'' into an operationally verified
rule-level
description and composition protocol, and we measured where it works and where it breaks. Under a
frozen shared symbolizer, eight independently trained agents can be described by comparable rule
banks anchored to replay-verified, hash-bound traces; offline fusion of those banks exceeded the
post-hoc-selected single-actor comparator on CartPole-v1 by $+42.31$ return points (95\% interval
$[23.98,61.93]$, exploratory because the comparator was chosen by training return and the task
exhibits ceiling effects) while
remaining auditable at the artifact level (replay and provenance checks pass under the evaluated
setup). The same protocol fails on Acrobot-v1 in a way that isolates the
fragile link---confidence-ranked arbitration is anti-selective under the evaluated conflict and
occupancy regime, and
uniformly random arbitration outperforms it. Rule overlap does not transfer to behavioral
agreement, and a fitted-Q GPI baseline collapses in both environments. The failure-driven design
history (lossy grammar compression, training-constraint interference, confidence-threshold collapse,
serialization and ledger-schema defects) is reported as evidence about which intuitions are unsafe.
The honest summary is bounded: rule-level auditing is operationally verified under the evaluated
replay, hashing, and provenance checks and is useful in low-conflict regimes;
rule-level composition is environment-dependent; and the boundary conditions we identified are the
paper's main deliverable. Within the scope evaluated here, the rule-level audit layer is a
verification substrate for composition rather than a claim that the present protocol already
generates novel skills.

\bibliographystyle{unsrt}
\bibliography{merged_main}

@article{williams1992simple,
  author  = {Williams, Ronald J.},
  title   = {Simple Statistical Gradient-Following Algorithms for Connectionist Reinforcement Learning},
  journal = {Machine Learning},
  volume  = {8},
  number  = {3-4},
  pages   = {229--256},
  year    = {1992}
}

@book{sutton2018reinforcement,
  author    = {Sutton, Richard S. and Barto, Andrew G.},
  title     = {Reinforcement Learning: An Introduction},
  edition   = {2nd},
  publisher = {MIT Press},
  year      = {2018}
}

@article{nevillmanning1997identifying,
  author  = {Nevill-Manning, Craig G. and Witten, Ian H.},
  title   = {Identifying Hierarchical Structure in Sequences: A Linear-Time Algorithm},
  journal = {Journal of Artificial Intelligence Research},
  volume  = {7},
  pages   = {67--82},
  year    = {1997}
}

@inproceedings{le2019batch,
  author    = {Le, Hoang and Voloshin, Cameron and Yue, Yisong},
  title     = {Batch Policy Learning under Constraints},
  booktitle = {Proceedings of the 36th International Conference on Machine Learning},
  year      = {2019}
}

@inproceedings{voloshin2021empirical,
  author    = {Voloshin, Cameron and Le, Hoang and Jiang, Nan and Yue, Yisong},
  title     = {Empirical Study of Off-Policy Policy Evaluation for Reinforcement Learning},
  booktitle = {NeurIPS Datasets and Benchmarks Track},
  year      = {2021}
}

@misc{towers2024gymnasium,
  author       = {Towers, Mark and others},
  title        = {Gymnasium: A Standard Interface for Reinforcement Learning Environments},
  howpublished = {arXiv:2407.17032},
  year         = {2024}
}

@misc{liu2025aim,
  author       = {Liu, Hung Ming},
  title        = {{AI} Mother Tongue: Self-Emergent Communication in {MARL} via Endogenous Symbol Systems},
  howpublished = {arXiv:2507.10566},
  year         = {2025}
}

@inproceedings{vandenoord2017neural,
  author    = {van den Oord, Aaron and Vinyals, Oriol and Kavukcuoglu, Koray},
  title     = {Neural Discrete Representation Learning},
  booktitle = {Advances in Neural Information Processing Systems 30},
  year      = {2017}
}

@inproceedings{lowe2019,
  author    = {Lowe, Ryan and Foerster, Jakob and Boureau, Y-Lan and Pineau, Joelle and Dauphin, Yann},
  title     = {On the Pitfalls of Measuring Emergent Communication},
  booktitle = {Proceedings of the 18th International Conference on Autonomous Agents and MultiAgent Systems (AAMAS)},
  pages     = {693--701},
  year      = {2019}
}

@inproceedings{foerster2016,
  author    = {Foerster, Jakob and Assael, Yannis M. and de Freitas, Nando and Whiteson, Shimon},
  title     = {Learning to Communicate with Deep Multi-Agent Reinforcement Learning},
  booktitle = {Advances in Neural Information Processing Systems 29},
  year      = {2016}
}

@inproceedings{havrylov2017,
  author    = {Havrylov, Serhii and Titov, Ivan},
  title     = {Emergence of Language with Multi-Agent Games: Learning to Communicate with Sequences of Symbols},
  booktitle = {Advances in Neural Information Processing Systems 30},
  pages     = {2149--2159},
  year      = {2017}
}

@inproceedings{bastani2018viper,
  author    = {Bastani, Osbert and Pu, Yewen and Solar-Lezama, Armando},
  title     = {Verifiable Reinforcement Learning via Policy Extraction},
  booktitle = {Advances in Neural Information Processing Systems 31},
  year      = {2018}
}

@misc{roth2019conservative,
  author       = {Roth, Aaron M. and Topin, Nicholay and Jamshidi, Pooyan and Veloso, Manuela},
  title        = {Conservative {Q}-Improvement: Reinforcement Learning for an Interpretable Decision-Tree Policy},
  howpublished = {arXiv:1907.01180},
  year         = {2019}
}

@inproceedings{barreto2017successor,
  author    = {Barreto, Andr{\'e} and Dabney, Will and Munos, R{\'e}mi and Hunt, Jonathan J. and Schaul, Tom and van Hasselt, Hado P. and Silver, David},
  title     = {Successor Features for Transfer in Reinforcement Learning},
  booktitle = {Advances in Neural Information Processing Systems 30},
  year      = {2017}
}

@inproceedings{henderson2018deep,
  author    = {Henderson, Peter and Islam, Riashat and Bachman, Philip and Pineau, Joelle and Precup, Doina and Meger, David},
  title     = {Deep Reinforcement Learning That Matters},
  booktitle = {Proceedings of the AAAI Conference on Artificial Intelligence},
  volume    = {32(1)},
  year      = {2018}
}

@inproceedings{fujimoto2019offpolicy,
  author    = {Fujimoto, Scott and Meger, David and Precup, Doina},
  title     = {Off-Policy Deep Reinforcement Learning without Exploration},
  booktitle = {Proceedings of the 36th International Conference on Machine Learning},
  pages     = {2052--2062},
  year      = {2019}
}

@inproceedings{kumar2020cql,
  author    = {Kumar, Aviral and Zhou, Aurick and Tucker, George and Levine, Sergey},
  title     = {Conservative {Q}-Learning for Offline Reinforcement Learning},
  booktitle = {Advances in Neural Information Processing Systems 33},
  year      = {2020}
}

@inproceedings{madumal2020explainable,
  author    = {Madumal, Prashan and Miller, Tim and Sonenberg, Liz and Vetere, Frank},
  title     = {Explainable Reinforcement Learning through a Causal Lens},
  booktitle = {Proceedings of the AAAI Conference on Artificial Intelligence},
  volume    = {34(03)},
  pages     = {2493--2500},
  year      = {2020}
}

@article{stappert2026integrating,
  author  = {Stappert, Mirko and Lutz, Bernhard and Goby, Niklas and Neumann, Dirk},
  title   = {Integrating Human Knowledge through Action Masking in Reinforcement Learning for Operations Research},
  journal = {Computers \& Industrial Engineering},
  volume  = {222},
  pages   = {112340},
  year    = {2026},
  note    = {arXiv:2504.02662}
}

@inproceedings{huang2022masking,
  author    = {Huang, Shengyi and Onta{\~n}{\'o}n, Santiago},
  title     = {A Closer Look at Invalid Action Masking in Policy Gradient Algorithms},
  booktitle = {Proceedings of the International Florida Artificial Intelligence Research Society Conference (FLAIRS)},
  volume    = {35},
  year      = {2022},
  doi       = {10.32473/flairs.v35i.130584},
  note      = {arXiv:2006.14171}
}

@article{garcia2015comprehensive,
  author  = {Garc{\'i}a, Javier and Fern{\'a}ndez, Fernando},
  title   = {A Comprehensive Survey on Safe Reinforcement Learning},
  journal = {Journal of Machine Learning Research},
  volume  = {16},
  number  = {1},
  pages   = {1839--1886},
  year    = {2015}
}

@misc{liu2026auditablerl,
  author       = {Liu, Hung Ming},
  title        = {{AIM} Auditability and Rule-Level {RL} Fusion: Full Training Traces and Checkpoints},
  howpublished = {figshare, dataset, version 1},
  year         = {2026},
  note         = {doi:10.6084/m9.figshare.33970177. Published 2026-09-23},
  doi          = {10.6084/m9.figshare.33970177},
  url          = {https://doi.org/10.6084/m9.figshare.33970177}
}

\appendix
\clearpage
\section{Per-Seed AIM Causal-Intervention Appendix}
\label{app:aim}

Table~\ref{tab:aim} reports the matched-token causal intervention per seed rather than only in
aggregate, so that seed-level variation is visible; it is provided in the spirit of not relying on a
single aggregate number. Each row is one independent run of the designed private-target communication
game (Section~\ref{sec:experiments:causal}). ``Active position'' is the message token whose parity
mutual information is maximal; $\Delta P(B\mid C)$ is the \emph{whole-message} change in the receiver's
action probability, whereas the matched-token contrast quoted in Section~\ref{sec:experiments:causal}
is slightly smaller (six-run mean $0.99837$, range $0.996986$--$0.999685$); ``shuffled'' is receiver
accuracy under a shuffled message; and ``MI'' is the
active-token mutual information in bits against a permutation null. Values are recomputed from the
archived run records of the six runs and are rounded for display.

\begin{table}[H]
\centering
\caption{Per-seed matched-token causal intervention on the designed private-target game. Reported to
expose seed-level variation. The scope is the designed task only and does not transfer to
state--action rule fusion.}
\label{tab:aim}
\begin{tabular}{lcccc}
\toprule
Seed & Active position & $\Delta P(B\mid C)$ (whole message) & Shuffled acc. & Active-token MI (bits) \\
\midrule
0 & 1 & 0.9958 & 0.522 & 0.985 \\
1 & 1 & 0.9977 & 0.503 & 0.992 \\
2 & 0 & 0.9985 & 0.507 & 0.996 \\
3 & 1 & 0.9961 & 0.496 & 0.994 \\
4 & 0 & 0.9997 & 0.502 & 0.999 \\
5 & 1 & 0.9994 & 0.491 & 1.000 \\
\midrule
Mean & --- & 0.9979 & 0.504 & 0.994 \\
\bottomrule
\end{tabular}
\end{table}

The per-seed rows show that the aggregate is not carried by a single run: every seed attains
$\Delta P(B\mid C)\ge 0.995$ with a shuffled-message control near chance ($0.49$--$0.52$) and an
active-token MI of $0.98$--$1.00$ bits. The active token position is not fixed across seeds
(position~1 for four runs, position~0 for two), which is why the intervention is reported per position
rather than as a single whole-sequence statistic.

\end{document}